\documentclass[runningheads]{llncs}
\usepackage{color}
\usepackage{capt-of}
\usepackage{amsmath,amssymb} 
\usepackage[width=122mm,left=12mm,paperwidth=146mm,height=193mm,top=12mm,paperheight=217mm]{geometry}
\usepackage{epsfig}
\usepackage{commath}
\usepackage{multirow}
\usepackage{makecell}
\usepackage{booktabs}
\usepackage{varwidth}

\newcommand{\etal}{\textit{et al}.}
\newcommand{\PMI}{\mbox{PMI}}

\usepackage{lipsum}

\newcommand\blfootnote[1]{%
  \begingroup
  \renewcommand\thefootnote{}\footnote{#1}%
  \addtocounter{footnote}{-1}%
  \endgroup
}

\begin{document}
\title{Concept Mask: Large-Scale Segmentation from Semantic Concepts}
\titlerunning{Large-Scale Segmentation from Semantic Concepts}
\authorrunning{Y. Wang, Z. Lin, X. Shen, J, Zhang, and S. Cohen}

\author{Yufei Wang\inst{1*}  \and
Zhe Lin\inst{2} \and
Xiaohui Shen \inst{3} \inst{*} \and
Jianming Zhang\inst{2} \and
Scott Cohen\inst{2}}

\institute{University of California, San Diego, CA, USA\\
\email{yuw176@ucsd.edu} \and
Adobe Research, San Jose, CA, USA \\
\email{\{zlin,jianmzha,scohen\}@adobe.com} \and
ByteDance AI Lab, Menlo Park, CA, USA \\
\email{shenxiaohui@bytedance.com}
}

\maketitle              

\begin{abstract}\blfootnote{$^*$ The work was done when the authors were in Adobe Research.}Existing works on semantic segmentation typically consider a small number of labels, ranging from tens to a few hundreds. With a large number of labels, training and evaluation of such task become extremely challenging due to correlation between labels and lack of datasets with complete annotations. 
We formulate semantic segmentation as a problem of image segmentation given a semantic concept, and propose a novel system which can potentially handle an unlimited number of concepts, including objects, parts, stuff, and attributes. We achieve this using a weakly and semi-supervised framework leveraging multiple datasets with different levels of supervision. We first train a deep neural network on a 6M stock image dataset with only image-level labels to learn visual-semantic embedding on 18K concepts. Then, we refine and extend the embedding network to predict an attention map, using a curated dataset with bounding box annotations on 750 concepts. 
Finally, we train an attention-driven class agnostic segmentation network using an 80-category fully annotated dataset. We perform extensive experiments to validate that the proposed system performs competitively to the state of the art on fully supervised concepts, and is capable of producing accurate segmentations for weakly learned and unseen concepts.
\keywords{semantic segmentation, large-scale segmentation, semi-supervised learning, weakly-supervised learning, zero-shot learning}
\end{abstract}

\section{Introduction}

Image segmentation has attracted a lot of attention in the recent years, and has achieved great progress with the success of Deep Neural Networks (DNN) \cite{FCN,noh2015learning,segnet,deeplab}. Two popular tasks of segmentation problems are semantic segmentation and instance segmentation. Existing semantic segmentation or scene parsing methods mostly consider a small number of classes and their extension to a large number of classes is challenging. The main difficulty comes from arising overlap between labels when the number of labels significantly increases: for the large-scale setting, labels at different levels or different branches in the WordNet hierarchy could have complex spatial correlations and subsequently confuse the pixel level annotation tasks. For example, for the face of a person, both the fine level annotation of ``face" and the higher level annotation of ``person" are correct, and for the area of ``clothing" on a human body can also be annotated as ``person" or ``body". This will cause a substantial challenge in training and evaluation of segmentation algorithms. On the other hand, pixel wise annotation for a large number of images and labels takes a lot of manual effort and is costly to obtain, and current publicly available benchmark datasets only have a small number of classes (for example, the MIT Scene Parsing Benchmark, the largest scene parsing dataset, contains annotations for only 150 classes). 

As for instance segmentation, the problem only focuses on objects, and the state-of-the-art bounding box proposal-based methods \cite{fcis,mnc,mask-rcnn} cannot handle object parts, stuff or other concepts like visual attributes. This is because the region proposal network predicts objectness of a bounding box, and naturally takes object parts or stuff as negative examples.

\begin{figure}[t]
\includegraphics[width=0.8\textwidth]{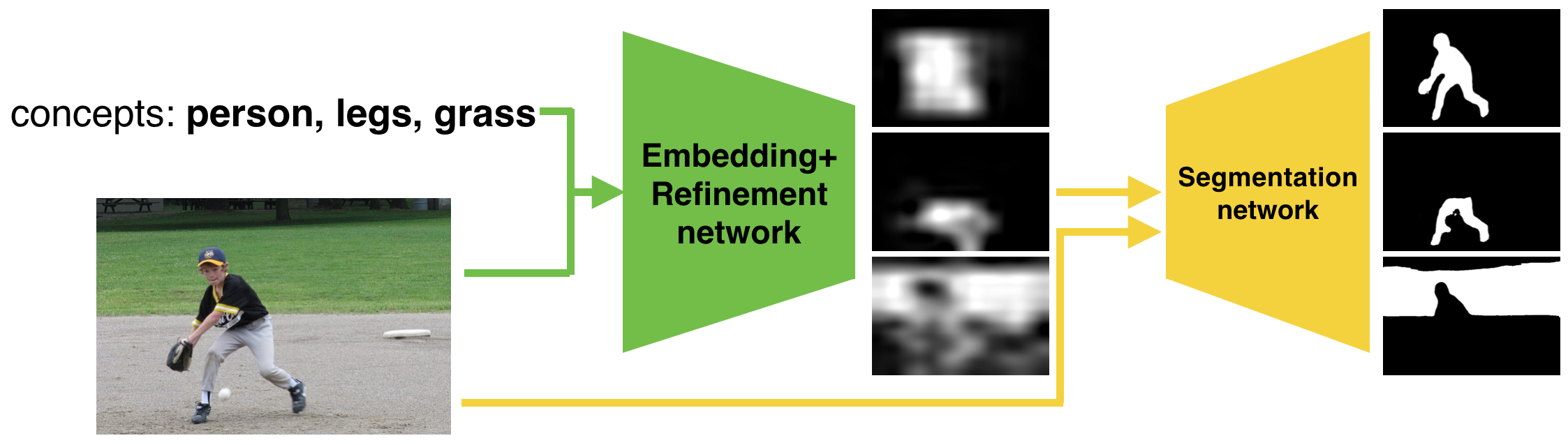}
\centering
\caption{Overall architecture of the proposed framework. Given a concept (can be object, object parts, stuff, etc.) and an input image, our embedding network and the subsequent attention refinement network predict a low resolution attention map, and a label agnostic segmentation network takes the attention map and the original image as input to predict a segmentation mask for the concept.}
\label{framework}
\end{figure}

In this work, we take a step forward and propose a new approach for large-scale semantic segmentation. To overcome the label ambiguity issue, we formulate the task as a problem of image segmentation given an arbitrary semantic concept. For example, the concept can refer to an object, object part, object group, stuff, attribute, etc. By this formulation, we alleviate the issue of label confusion in large-scale semantic segmentation and scene parsing, which makes training and evaluation of segmentation algorithms more well-defined.

However, there is no available dataset for large-scale segmentation. To leverage the existing datasets with different levels of supervision, we use four datasets for training: a 6M Stock dataset (crawled from a stock website) with 18K image level labels; a curated a 750-concept dataset from Open Images \cite{openimages} and Visual Genome \cite{vg}, with bounding box annotation; MS-COCO \cite{coco} with full segmentation annotation for 80 object classes. In order to evaluate the model's capability on weakly supervised learning, we select a diverse set of 50 test concepts among 18K concepts excluding those 750 concepts with the bounding box annotations. 

Given the datasets, we propose a new weakly and semi-supervised learning approach which can leverage all the available training data in an incremental learning framework. The proposed incremental learning framework consists of three steps. First, we train a deep neural network on the stock dataset\footnote{\url{https://stock.adobe.com}} to learn large-scale visual-semantic embedding between images and 18K concepts \cite{zhang2016EB}. By running the embedding network in a fully convolutional manner, we can compute a coarse attention (heat) map for any given concept. Next, we attach two fully connected layers to the embedding network and fine-tune the refinement network in low resolution using the 750-concept dataset with bounding box annotations to obtain improved attention maps. We use multi-task training to learn from the new 750-concept supervision without affecting the previously learned knowledge on 18K concepts. Finally, we train a label-agnostic segmentation network which takes the attention map and original image as input and predicts a high-resolution segmentation mask without much knowledge of the concept of interest. The segmentation network is trained with only 80 object categories with pixel-level supervision but we show that it generalizes well to any semantic concept, including objects, object parts, and even background stuff, due to the use of attention maps for class-agnostic segmentation.\footnote{Note that traditional bounding box proposal-based methods with class-agnostic segmentation could easily fail to detect proposals on object parts or stuff and the bounding box-based class-agnostic segmentation module trained with object categories cannot deal with stuff categories.} During testing, we can attach the segmentation network to the attention network to form a unified feed-forward network model.

We perform extensive experiments to validate that the proposed approach performs competitively to the state of the art on fully supervised concepts, and is capable of producing accurate segmentations for weakly learned and unseen concepts. The main contributions of this paper are as follows: 1) We address the problem of large-scale semantic segmentation with a new formulation: large-scale segmentation given a concept; 2) To study this task, we construct multiple datasets with different levels of supervision, and establish performance evaluation methods; 3) We propose a new, incremental learning approach that can predict segmentation masks for a very large number of concepts, including object, object parts, and stuff; 4) We propose a novel auxiliary loss called spatial discrimination loss for discriminative segmentation training for a large number of concepts with complex semantic relationships.

\section{Related Work}
\subsubsection{Fully Supervised Semantic Segmentation}
Semantic segmentation has made remarkable progress with the recent advancement in deep convolutional neural networks (CNN). Many CNN-based segmentation networks \cite{psp,FCN,mnc,mask-rcnn,fcis,city} perform well on datasets with a small number of labels, such as PASCAL VOC \cite{pascal} with 20 object classes, ADE20K\cite{ade} with 150 stuff/object classes. 

For instance aware segmentation which requires segmentation of individual object instances, methods based on region proposals \cite{rcnn} perform well on the COCO dataset with 80 object classes \cite{fcis,mnc,mask-rcnn}. However, the region proposal based methods can only handle object classes, and their generation to other concepts such as object parts or stuff is not straightforward.


These methods are all fully supervised, and assume disjoint classes, which enables training segmentation networks with a discriminative soft-max loss.

\subsubsection{Weakly/Semi-supervised Semantic Segmentation}
In order to reduce annotation efforts needed in fully supervised methods, weakly supervised segmentation methods have been proposed~\cite{weak_add1,weak_add2,weak_add3,web,weak1,weak2,weak3}. Image-level annotations require minimum manual effort, but methods with such annotations have a large performance gap compared to fully supervised methods; additional label types such as bounding box annotations are exploited to improve the performance. On the other hand, some works exploit complementary data from the web \cite{web}. Those weakly supervised methods still focus on a small set of disjoint labels.

Different from those works, this paper aims to scale semantic segmentation to a very large number of categories. We make use of all the available annotations in several datasets, thus combining different levels of annotation.

One work related to our model is by Hong \etal~\cite{decouple}. Segmentation is decoupled into two tasks with two separate networks: classification and segmentation. The classification network uses image level annotation, and the segmentation network uses pixel level annotation. However, their work still focuse on a very small number of labels, and their model cannot generalize to unseen concepts.

Another recent work related to ours is by Hu \etal~\cite{everything}. It aims at instance segmentation on a large number of categories with a small fraction of mask annotations and a large fraction of box annotations. In contrast, our work aims to segment not only objects, but also other concepts such as stuff, parts, and  visual attributes (like color); our model is learned to segment concepts trained with only image-level supervisions and can even handle unseen concepts.


\subsubsection{Zeroshot Learning}
For the problem of zeroshot learning, models are tested on unseen categories by transferring knowledge from the trained categories. Semantic embedding of vectors associated with class labels are obtained from object attribute labels \cite{zeroshot1,zero2,zero3} or word embeddings learned from linguistic tasks \cite{devise,NIPS2013_5027,zero4}.  Zeroshot learning can also be applied to segmentation tasks. With the embedding network that maps an image to a word embedding space, segmentation models have the potential to generate masks given an unseen concept~\cite{open_vocab}.

\subsubsection{Large Scale Segmentation/Parsing}
Zhao \etal~aim to recognize and segment objects with open vocabulary \cite{open_vocab}, which is in line with our goal of large scale segmentation. Words and images are embedded into a joint space to allow zero-shot learning. Our work is different from theirs in that (1) Zhao \etal~address only zero-shot segmentation while we aim to solve weakly supervised and zero-shot segmentation in a unified  framework, (2) Zhao \etal~use WordNet for modeling hierarchical label relationships while we consider more complex label relationships including spatial overlap/exclusion which makes trained segmentation models more discriminative, (3) Zhao \etal~view the open-vocabulary scene parsing as a concept retrieval problem, whereas we assume a target concept is given as an extra input to predict the segmentation mask. Our task is easier to evaluate and with less ambiguity in the ground truth masks. 

\section{Dataset}
Public segmentation datasets typically contain pixel-level annotations on only a small number of labels. On the other hand, datasets with a much larger vocabulary are only weakly annotated, either with bounding box or image-level labels. To make the most of the available datasets, we form a  combined dataset, containing different levels of annotations:

\begin{itemize}
  \item \textbf{COCO-80}: MS-COCO dataset~\cite{coco} pixel level annotation on 80 categories. 
  \item \textbf{OIVG-750}: Combined Open Images~\cite{openimages} and Visual Genome~\cite{vg} dataset with 750 concepts (including COCO concepts) with bounding boxes.\footnote{Visual Genome has more than 10000 classes, and Open Images has 545 trainable classes. We merge the labels of the two datasets, and filtered out classes with very few examples. 750 concepts containing objects, object parts, and stuff are selected.} 
  \item \textbf{Stock-18K}: 6M Stock dataset annotated with 18K tags. 
\end{itemize}

With the combined dataset, we can evaluate the performance of segmentation methods under different levels of supervision by constructing the following test set: (1) strongly supervised concepts: COCO-80 test set, (2) box-level weakly supervised concepts: Weak-Box-670, obtained by excluding 80 COCO categories from OIVG-750, (3) image-level weakly supervised concepts: Weak-Image-50, obtained by choosing 50 new concepts from OIVG excluding OIVG-750. 

For testing on weakly-supervised settings, there is no available segmentation ground truth for classes outside COCO-80. Therefore we generate pseudo-ground truth from bounding boxes, using an automatic segmentation model \cite{xuning}. It is then manually cleaned up, and in the supplementary material we show some examples of the generated pseudo-ground truth masks. Note that we use the pseudo-ground truth masks only for evaluation. 

\section{Proposed Approach}
\begin{figure}[t]
\includegraphics[width=1.\textwidth]{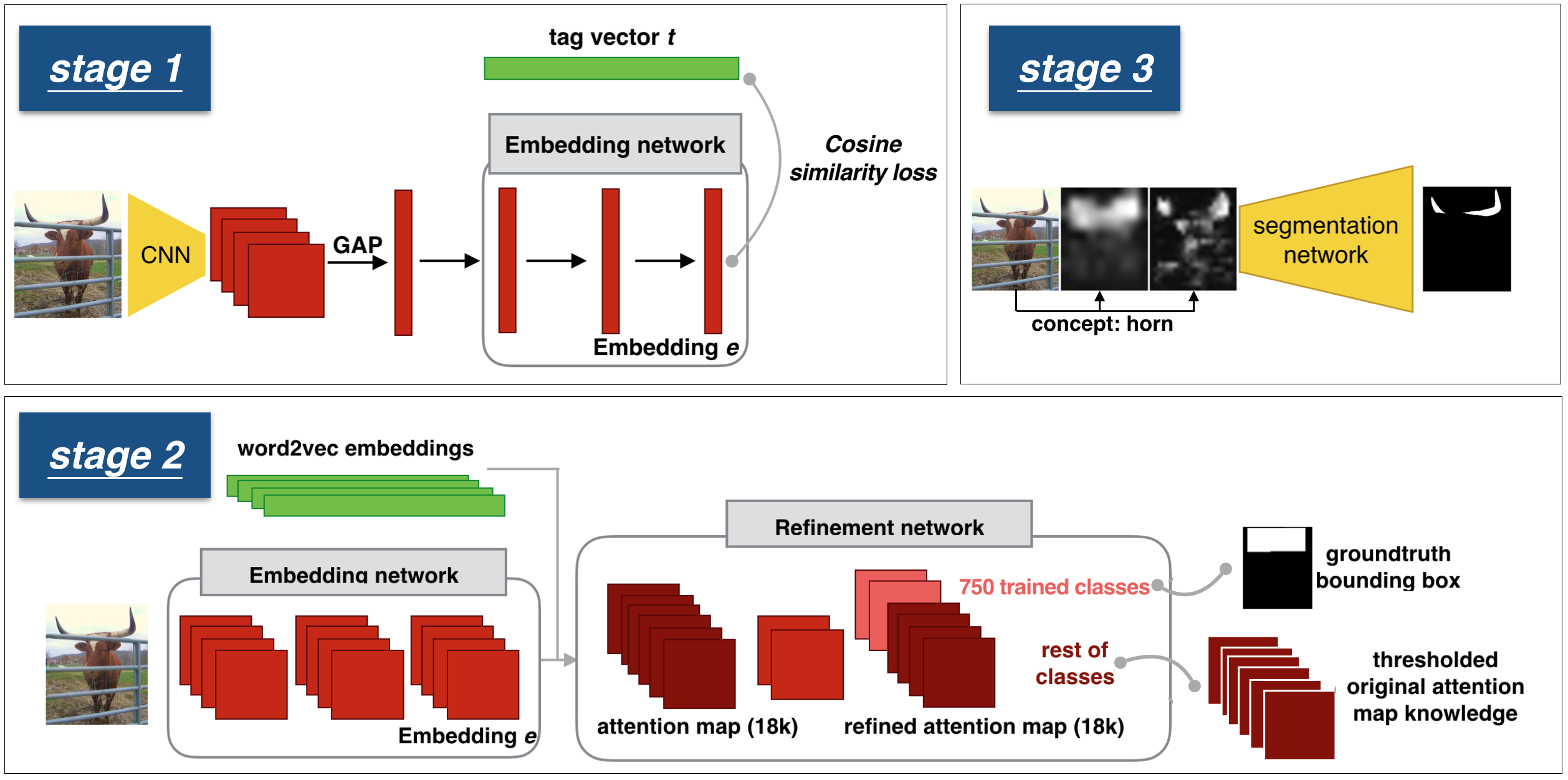}
\centering
\caption{Three stages of the training framework. Stage1: embedding network trained on image level annotation. Stage2: multi-task training of attention network. It finetunes the embedding network while training the refinement network from scratch. It refines the attention map on 750 concepts with bounding box supervision, meanwhile preserves the knowledge learned from embedding network on 18K concepts. Stage3: label agnostic segmentation network that takes the original image and two refined attention maps (generated from input image with two scales), and predicts the segmentation mask.
}
\label{step123}
\end{figure}

The overall framework of our large-scale segmentation system is illustrated in Figure~\ref{framework}. It is composed of an embedding and refinement network that produces an attention map from the input image and a specified concept, and an attention-driven label-agnostic segmentation network that predicts a final segmentation mask. Utilizing three different 
levels of supervision, we train the entire framework incrementally in three stages: 
\begin{enumerate}
  \item Train an embedding network on Stock-18K that learns the visual-semantic embedding between images and 18K semantic concepts. Only image-level annotations are used in this stage. After training, we transform the network to fully convolutional, which can generate a low resolution attention map given an input image and any of the 18K concepts.
  \item Append a refinement module to the end of the embedding network, and train the refinement network on OIVG-750 with bounding box annotations, aiming at obtaining attention maps of higher quality.
  \item Train a label agnostic segmentation network on COCO-80 with 80-class full segmentation supervision. The network takes the initial attention maps together with the image as input, and predicts a higher resolution segmentation mask with more accurate boundaries for the concept.
\end{enumerate}

\subsection {Embedding Network}
We first utilize the Stock-18K dataset with image-level annotations to learn large-scale visual-semantic embedding. The dataset has 6 million images, each with heavily annotated tags from an 18K vocabulary. The training set is denoted as $\mathcal{D} = \{ (I, (w_1, w_2, ..., w_n) \}$, where $I$ is an image and $w_i$ is the word vector representation of its associated ground-truth tags.

\subsubsection{Word Embedding}
Instead of using off-the-shelf word embeddings trained on a text corpus, we use point-wise mutual information (PMI) to learn our own word embeddings for each tag $w$ in the vocabulary. PMI is a measure of association commonly used in information theory and statistics \cite{pmi}. We follow \cite{pmi2} to calculate PMI matrix and then do eigenvector decomposition to the matrix to get the word vector. More details are shown in the supplementary material. 


Since each image is associated with multiple tags, in order to obtain a single word vector representation of each, we calculate a weighted average over all the associated tags: $t = \sum_{i=1}^n \alpha_i w_i$ where $\alpha_i = -\textup{log}(p(w_i))$ is the inverse document frequency (idf) of the word $w_i$. We call the weighted average \textit{soft topic embedding}.

\subsubsection{Joint Word-Image Embedding}
The embedding network is learned to map the image representation and the word vector representation of its associated tags into a common embedding space. As shown in Figure~\ref{step123} stage 1, each image $I$ is passed through a CNN feature extractor. Here we use ResNet-50 \cite{resnet} as feature extraction network. After global average pooling (GAP), the visual feature is then fed into a 3-layer fully connected network, denoted as \textbf{Embedding network}, with each fc-layer followed by a batch normalization layer and a ReLU layer. The output is the visual embedding $e = embed\_net(I)$, and is align with the soft topic word vector $t$ by a cosine similarity loss:
$L_{embed}(e,t) = 1 - \frac{e^T t}{\norm{e} \norm{t}}$.

\subsubsection{Attention Map}
After the embedding network is trained, to predict an attention map for a given concept, we remove the global average pooling layer, and transform the network to a fully-convolutional network by converting the fully connected weights to $1 \times 1 $ convolution kernels and the batch normalization layers to spatial batch normalization layers. After this transformation, we can obtain a dense embedding map given an image and a word vector, in which the value at each location is the similarity between the word and the image region around that location. Thus the embedding map can also be viewed as an attention map for that word. Note that the way we generate the attention map is similar to~\cite{zhou2016cvpr}. However, we use soft topic embedding instead of discriminative classification so the attention map has better spatial coverage than the one in~\cite{zhou2016cvpr}.

Formally, the attention map for a given concept $w$ can be calculated as:
\begin{equation}
\alpha_{(i,j)}^0 = <e_{i,j}, w>
\label{eqn:attention}
\end{equation}
where $(i,j)$ is the location index for the attention map. For an unseen concept that is not used in our image-word embedding training, as long as we can obtain its word vector $w$, we can still obtain its attention map using Eqn.\ref{eqn:attention}. Therefore, our embedding network can be generalized to any arbitrary concept.

\subsection {Attention Map Refinement}
Although the embedding network trained on image level annotation can predict attention maps for any given word vector, the quality of the attention maps is still very coarse due to the lack of annotations with spatial information.

In order to improve the quality of the attention map, we leverage existing finer-level annotations, namely the object bounding box annotations that are available in several large-scale datasets. Specifically, we use the  OIVG-750 dataset to train a network for attention map refinement.

\subsubsection{Refinement Network Architecture} As shown in Figure~\ref{step123} stage 2, the refinement network is appended at the end of the embedding network, and is composed of two convolutional layers with $1 \times 1$ kernels followed by a sigmoid layer. By treating word embeddings as convolutional kernels, embedding network can now output 18K coarse attention maps. The two-layer refinement network takes those coarse attention maps as input, and learns a non-linear combination of the concepts to generate refined attention maps for the 750 classes. This encourages the refinement network to consider relationships between concepts during training.

\subsubsection{Multi-task Training} For a given concept, training signal for its attention map is a binary mask based on the ground-truth bounding boxes, and a sigmoid cross entropy loss is used. Embedding network is also finetuned for better performance. However, since the bounding box annotations are only available for the 750 concepts, if we only train the network on those classes, the previously learned attention maps for the rest of 18K concepts will be corrupted if we also finetune the embedding network layers. Inspired by \cite{zhizhong16} on learning without forgetting, in order to preserve the learned knowledge from the rest of 18K concepts, an additional matching loss is added: the original attention maps generated by the embedding network are binarized with a threshold, and sigmoid cross entropy loss is exerted for the refined attention maps to match the original attention maps. The multi-task loss function is therefore as follows:

\begin{equation}
L = L_{xe}(G, \alpha) + c\sum_{k \in \Psi_N }L_{xe}(\textup{B}(\alpha_k^0), \alpha_k)
\label{eq1}
\end{equation}
where $L_{xe}(p, q)$ is the cross entropy loss between true distribution $p$ and predicted distribution $q$. $\alpha$ is the attention map of the given concept, $G$ is the ground truth mask with 1 being inside the bounding box, 0 outside. $\textup{B}(\alpha)$ is the binary mask after thresholding the attention map. $\alpha_k^0$ and $\alpha_k$ are original attention map and refined attention map respectively. $\Psi_N$ is the set of indices of top $N$ attention maps with the highest activation. The matching loss is exerted on attention maps with high activation only to avoid bias toward irrelevant concepts. $c$ is the weight balancing the losses. We choose $N = 800$, and $c = 10^{-6}$.

\subsubsection{Spatial Discrimination Loss} The reason we used sigmoid cross entropy loss during training instead of softmax loss as in semantic segmentation is that there are many concepts whose masks are overlapping with each other. It is especially common for objects and their parts. For example, the mask of face is always covered by the mask of person. Using softmax loss therefore would discourage the mask predictions on those concepts one way or another. At the same time, there are still many cases where the masks of two concepts never overlap. To utilize such spatial relationships between label pairs and make training of the attention maps more discriminative, we propose a novel auxiliary loss for discriminating those spatially non-overlapping concepts, referred to as spatial discriminative loss, to discourage high responses for spatially conflicting concepts occuring at the same time. 

In particular, we calculate the mask overlap ratio between every co-occurred concept pair in the training data:

\begin{equation}
O(i, j) = \frac{ \sum_n |a_n(i) \cap a_n(j)| }{\sum_n |a_n(i)|}
\end{equation}
where $a_n(i)$ is the mask of the $i$-th concept in image $n$, and $|a_n(i) \cap a_n(j)|$ is the overlapping area of between concepts $i$ and $j$. Here image $n$ has to include both $i$  and $j$ concepts to avoid impact of incomplete annotation. Note that the mask overlap ratio is non-symmetric. In the supplementary material, we show a subset of the overlap ratio matrix $O(i,j)$.

With the overlap ratio matrix $O(i,j)$, a training example of a concept $i$ can serve as a negative training example of its non-overlapping concept $j$, i.e., for a particular location in the image, the output for concept $j$ should be 0 if the ground-truth for concept $i$ is 1. To soften the constraint, we further weight the auxiliary loss based on the overlap ratio, where the weight $\gamma$ is calculated as:

\begin{equation}
\gamma_{ij} = \left\{\begin{matrix} 1- O(i, j), &   \mbox{if}\; \; O(i, j) < 0.5\\ 0, & \mbox{otherwise} \end{matrix}\right.
\end{equation}

\subsection {Label Agnostic Segmentation Network}

Our attention map refinement network now can predict low resolution attention map for an arbitrary concept using its word vector representation. To further obtain the mask of the concept with higher resolution and better boundary quality, we train a label agnostic segmentation network that takes the original image and the attention map as input, and generates a segmentation mask without knowing the concept, as shown in Figure~\ref{step123} stage 3. Since the goal of the segmentation network is to generate foreground segmentation mask given the prior knowledge of attention map, the segmentation network can generalize to unseen concepts, even though it is entirely trained on COCO-80 with only 80 object classes. 

To segment the masks at different scales, we generate multiple attention maps by feeding the embedding network with different input image sizes (300 and 700 in our experiments). The resultant attention maps are then upsampled to serve as the extra input channel to the segmentation network along with the image.

To make the segmentation network focus on generating accurate masks instead of having the extra burden of predicting the existence of the concept in the image, we normalize the attention maps to $[0,1]$. 
We found that such training strategy can learn better segmentation networks. During testing, the attention maps are normalized in the same way, and the verification of the existence of the concept is done separately, with details presented in the supplementary material.

For the architecture of segmentation network, we use an architecture that extracts and combines high level and low level features to predict a concept mask with accurate boundaries. See the supplementary materials for more details.

\subsection {Weakly Supervised Segmentation}
\label{zeroshot}
During testing stage, for the 18K concepts that are only trained with image level supervision, we do not directly use the attention map from the refinement network for that concept as  the input to the segmentation network. This is because the segmentation network only sees the examples of the COCO-80 during training, which has the attention map trained with bounding box / pixel-wise segmentation supervision. Thus, the discrepancy between the lower-quality attention maps of the 18K concepts and the higher-quality attention map of the 750 concepts will impact the segmentation performance on 18K concepts.

Therefore, for a concept $q$ from the 18K concepts with image level supervision, we find its nearest neighbor concept $p$ in the embedding space from the 750 concepts, and the attention maps is a linear combination $\alpha = \theta \alpha_q + (1-\theta) \alpha_p$ of the attention maps from the two concepts, with $\theta$  decided on validation set.


\section{Experiments}
In this section, we provide visual and numerical results on attention map prediction and segmentation mask generation under different levels of supervision. Experimental details are shown in the supplementary material.
\subsection{Datasets}
For COCO-80, we use the train2014 split, with 80k training images. For OIVG-750, there are 540k training images, and training examples for each concept varies from 8 to 100k. Stock-18K has 6M training images, and 30 tags for each image on average. Test/validation set for OIVG-750, Weak-Box-670, Weak-Image-50 have 5-10 examples per concept, and one example is held for validation.

\begin{figure}[!htb]
    \centering
    \begin{minipage}{0.49\textwidth}
        \centering
        \includegraphics[width=0.9\linewidth]{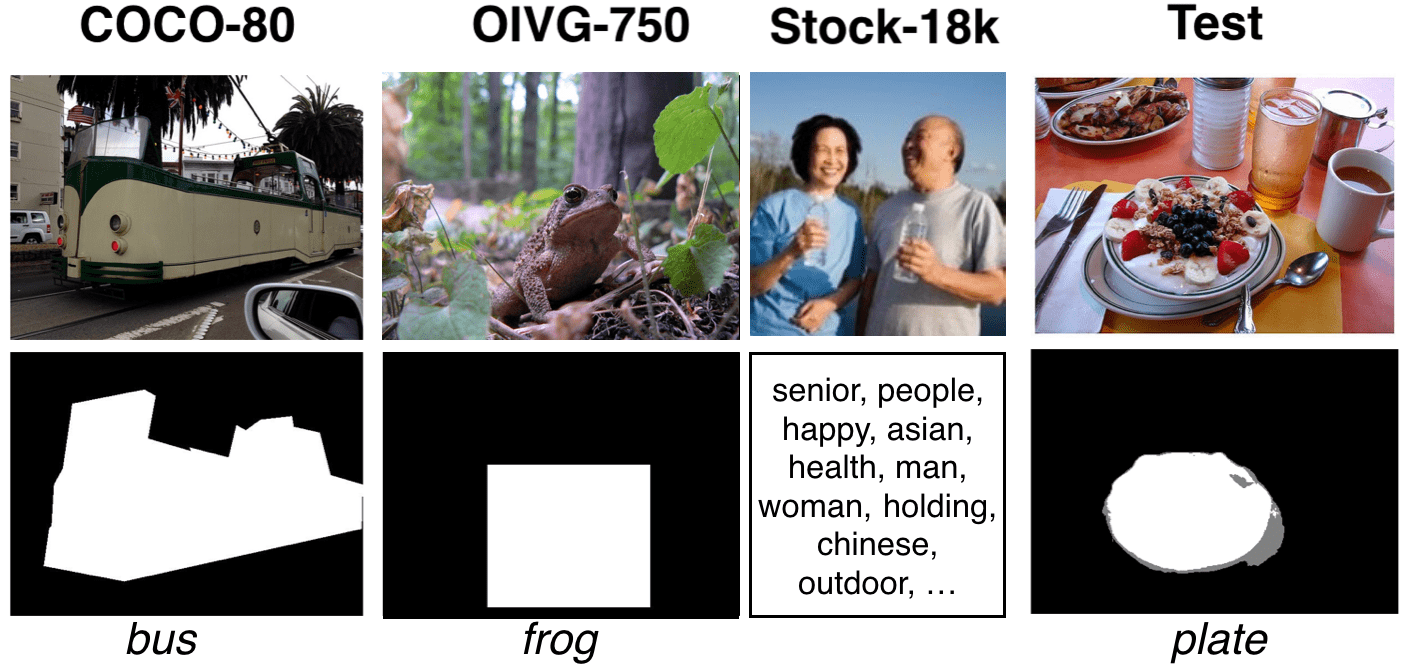}
        \captionof{figure}{Examples of our datasets with different levels of annotation. First row is the original image, and second row shows the annotation labels. The Stock-18k image is from \textbf{arekmalang - stock.adobe.com}.}
        \label{eg}
    \end{minipage}\hfill
     \begin{minipage}{0.49\textwidth}
    \centering
    \includegraphics[width=0.9\linewidth]{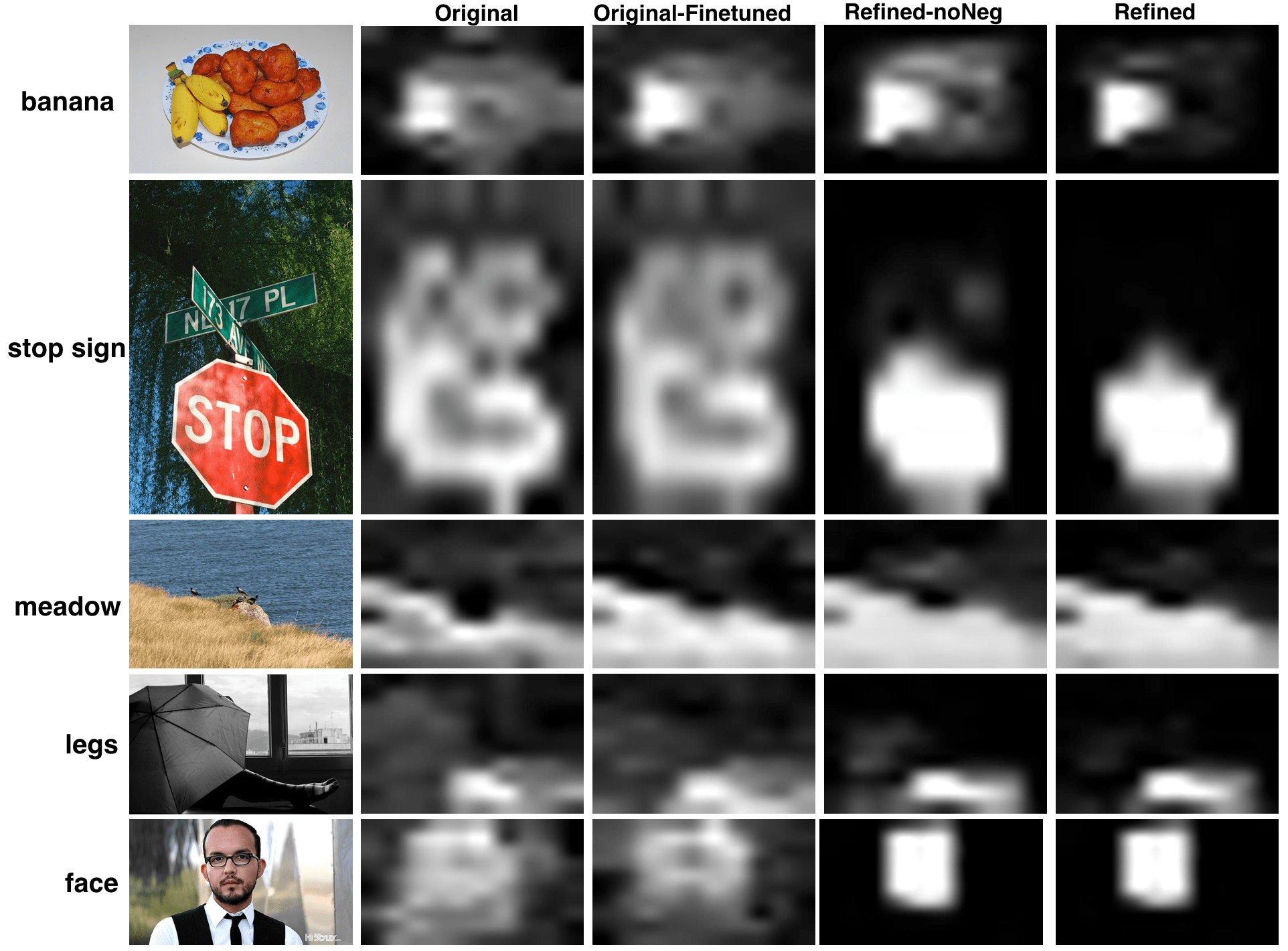}
    \captionof{figure}{Examples of attention map generated from different models/phases.}
    \label{result_heat}
  \end{minipage}

\end{figure}

In Figure~\ref{eg}, we show the example images and ground truth labels from different datasets. Each row shows an example of a dataset with annotation. In the last column, we also show an example test image from Weak-Image-50, for which the annotation is the pseudo-ground-truth mask.

\subsection{Attention Map Evaluation}
For attention map generation, we use several ways for evaluation. Following \cite{zhang2016EB}, we use Pointing Game for evaluation. For an image, if the maximum point in attention map lies in the ground truth mask, a hit is counted. We can measure the mean accuracy across all the concepts. We can also use IOU for evaluation. The attention map is a probability mask that ranges from 0 to 1, and we calculate IOU as follows:
\begin{equation}
\textup{IoU}_n = \frac{\alpha_n * G_n}{\textup{max}(\alpha_n, G_n)}
\end{equation}
where $n$ is image index, $\alpha$ is the attention map, and $G$ is the ground truth. When there is only bounding box ground truth available, we use the pseudo-ground-truth for evaluation. 

\begin{table}[ht]
    \centering
    \caption{Performance of different models on attention map generation}
\begin{tabular}{c|c|c|c|c}
\hline
              & Original & Original-Finetuned & Refined-noNeg & Refined        \\ \hline
Pointing Game & 0.578    & 0.631              & 0.806         & \textbf{0.810} \\ \hline
IoU           & 0.262    & 0.288              & 0.416         & \textbf{0.421} \\ \hline
\end{tabular}
    \label{table1}
 \end{table}

In Table~\ref{table1}, we compare the performance of different models/phases on attention map generation, using OIVG-750 evaluation set. \textbf{Original} is the original attention map we obtain from the embedding network, as described in Section 4.1. \textbf{Original-Finetuned} is the attention map from the embedding network, after finetuning with the refinement network. \textbf{Refined-noNeg} is the result from the refinement network as described in Section 4.2, without using the negative examples from non-overlapping concept. \textbf{Refined} is our full model. 

In Figure~\ref{result_heat}, we also show the visual result of attention map generated from different models. We can see that the original attention map already generates acceptable attention maps, but it is noisy, and sometimes locating to objects when the concept is object part (see example of \textit{face}). The refined attention map is much cleaner, and is covering the whole object/stuff. The comparison between the result from Refined-noNeg and Refined shows that by using negative samples from non-overlapping concept, the attention map is cleaner visually, and is more discriminative.

\subsection{Segmentation Evaluation}
In this section, we evaluate our model with different levels of supervision, and compare our results with different baselines.
For quantitative evaluation, we simply binarize the soft segmentation outputs using the threshold of 0.5 for all models. Given the binary ground truth mask and prediction mask for one concept, we can calculate precision/recall/IoU. Over different concepts, we calculate mean precision, recall, and mean IoU.

\subsubsection{Dataset}
Our model uses different levels of supervision, and thus we can evaluate the model performance on concepts with different levels of supervision separately: categories inside \textbf{COCO-80} are used to train the attention network and segmentation network, therefore we can evaluate it for full (strong) supervision. We use 5000 miniVal2014 split for evaluation, and all the annotated concepts in the images are evaluated; \textbf{Weak-Box-670} is used to evaluate segmentations with bounding box-level supervision; \textbf{Weak-Image-50} is used to evaluate our model's performance on concepts with only image-level supervision.

\subsubsection{Results}
We compare the performance of our model and baselines on different levels of supervision in Table~\ref{result2} and Table~\ref{result3}.

\begin{table}[h]
\caption{Comparison of the performance of our model and baselines, on different levels of supervision. On the left, we show the descriptions of different baselines we use}
\label{result2}
  \includegraphics[width=\linewidth]{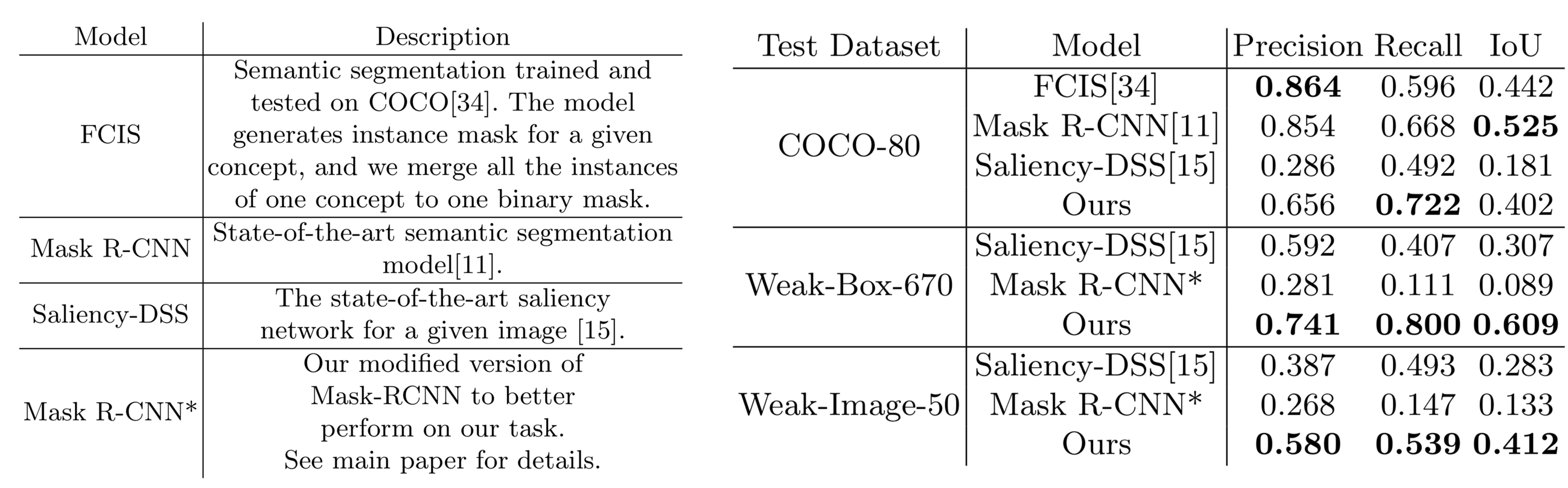}
\end{table}

\begin{table}[h]
\caption{Ablation study for our model on different levels of supervision. On the left we show the descriptions of the models we compare with}
\label{result3}
  \includegraphics[width=\linewidth]{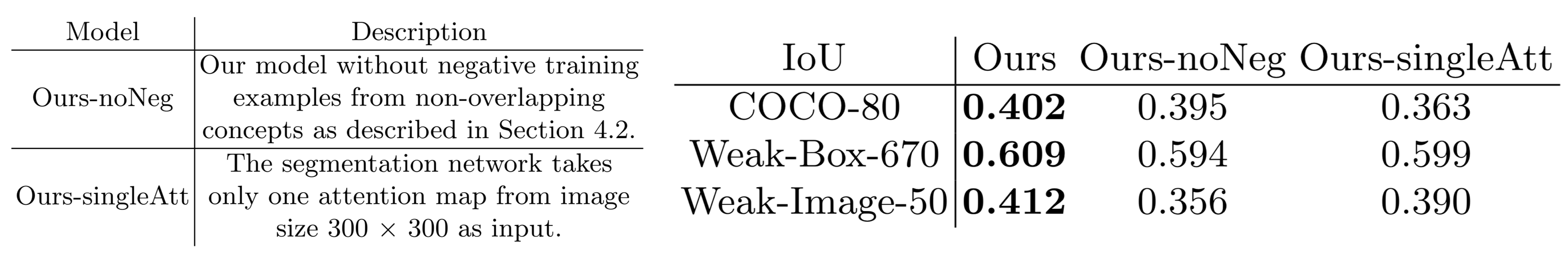}
\end{table}

As shown in Table~\ref{result2}, the performance of our model decreases with less supervision.  For the fully supervised concepts, our method performs competitively with semantic segmentation model FCIS, with 4\% of gap, and has moderate gap with curent state-of-the-art semantic segmentation model Mask R-CNN. The gap is predictable, because our model handles not only 80 categories inside COCO-80, but also orders of magnitude more concepts outside COCO. The closeness between the two models shows that although our model aims at a much bigger concept set, it performs very well on COCO concepts. 

For the weakly supervised setting, we first compare our method with the saliency baseline. Since the concepts for evaluation are manually picked and the images used for evaluation are manually filtered for insuring the quality of the groundtruth, one concern is that the test set is not sufficient to test concept segmentation, and a saliency object detection is enough. Here by showing the performance of the saliency detection model is poor, we demonstrate that our test dataset is valid for evaluating our task. 

We also train a modified Mask-RCNN (notated as Mask-RCNN*) for weakly supervised results. The original Mask R-CNN \cite{mask-rcnn} predicts a mask independently for each of the 80 COCO classes. For an RoI associated with ground-truth class $k$, loss is defined as per pixel sigmoid loss only on the $k$th class. However, this does not apply to our problem, because there is no mask annotation on our large scale OIVG-750 dataset. Therefore, we modify the segmentation head to predict a label-agnostic mask, which is trained only on COCO-80. Mask-RCNN* does not perform well, and the reason can be summarized as follows: First, Mask-RCNN cannot handle stuff classes, such as sky, tree, etc., because the segmentation head only sees 80 object classes with bounding boxes. In contrast, our two-stage model does not rely on bounding boxes, so it can handle stuff very well even though the segmentation head is only trained on 80 object classes. Second, for the large number of classes (750) and highly overlapping concepts, Mask-RCNN has a very low box detection rate, which might be due to conflict of bounding box proposals among object parts, stuff, object classes. 

For COCO-80, the IoU for the saliency detection result is very low, whereas for the other test dataset, the saliency performance is higher than that in COCO-80. This is due to the distributions of concepts in OIVG and COCO are essentially different, the former contains more cases with larger objects/stuff. Despite the higher performance on those test sets, we still see a great improvement of our method over saliency.

In Table~\ref{result3}, we show an ablation study of our approach with respect to the final performance. Our full model outperforms Ours-noNeg and Ours-singleAtt on all three cases of supervision setting, indicating the necessity of negative examples and the multi-scale attention map input to the segmentation network.

\begin{figure}[t]
\includegraphics[width=\textwidth]{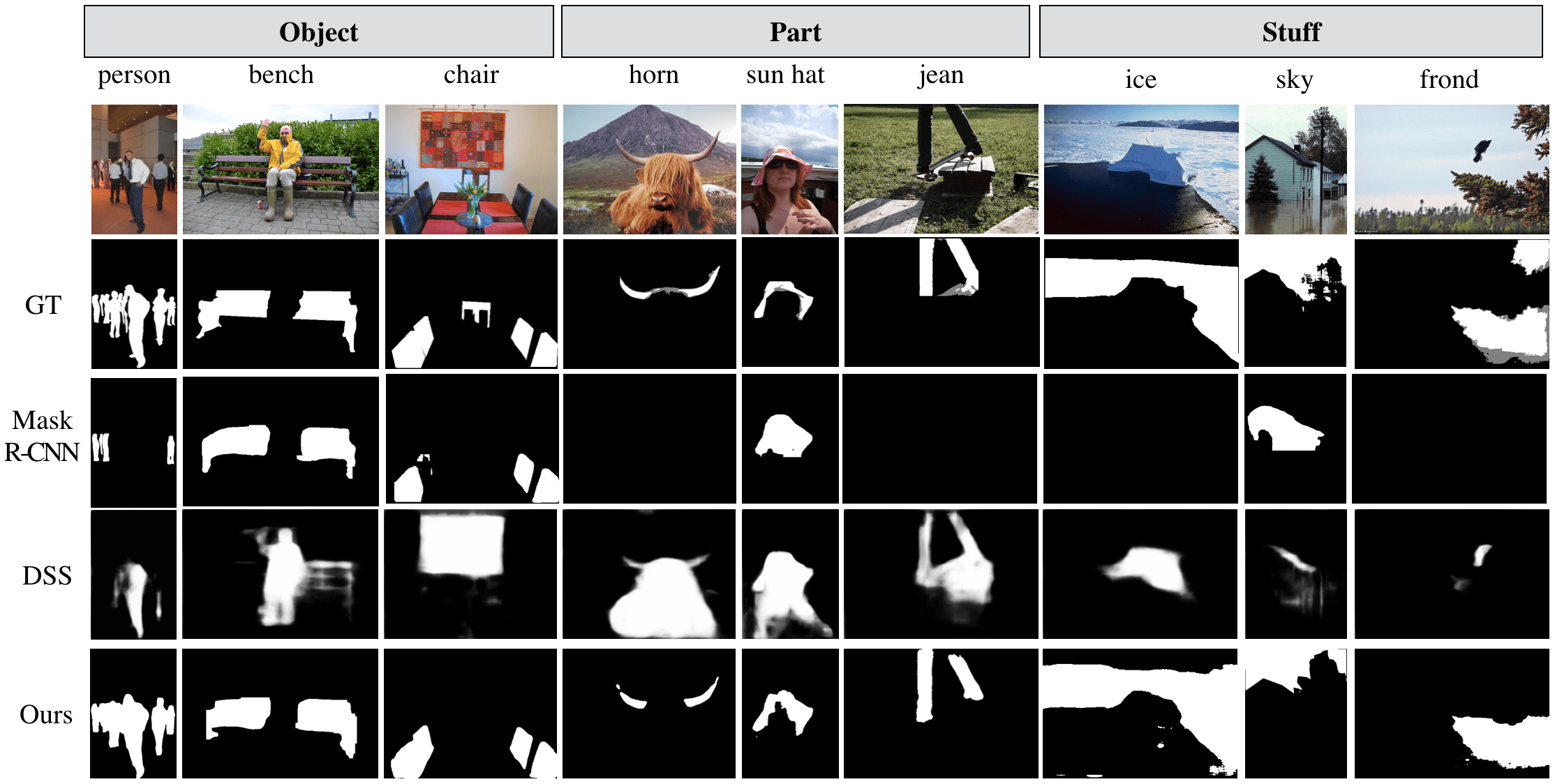}
\centering
\caption{Example of visual result of our segmentation network.}
\label{result-seg}

\end{figure}

Figure~\ref{result-seg} shows visual examples of our segmentation result. For object, object part, and stuff, we show three examples from each type. All three object categories are from COCO-80. \textit{jean} and \textit{frond} are from Weak-Image-50. The rest four categories are from Weak-Box-670. We also show different baselines' results. We provide more qualitative results and failure cases in the supplementary.

\subsection{Zero-shot Learning}
Since we use an embedding network for attention map prediction, with the word embeddings, our model can potentially handle unseen concepts. To test this potential, we curate 10 concepts outside the 18K concepts that our model is trained on, each with 5-10 test examples. The IoU of our method is 0.436, and the IoU of Saliency-DSS is 0.298. Further study with a larger test set is needed to fully justify the zero-shot learning ability of our model.

\section{Conclusion}
In this paper, we study semantic segmentation at a very large scale. 
With a large number of labels, training and evaluation of segmentation models are very challenging due to complex correlations between labels. To address the issue, we formulate the problem as conditional image segmentation given a semantic concept.
Under this formulation, we propose a powerful weakly and semi-supervised segmentation framework that can handle a large number of concepts including objects, parts, stuff, attributes, and even unseen concepts. The framework consists of three parts: 1) an embedding network that maps the image and a large scale of concepts into the same space; and 2) an attention network which refines the embedding network to predict low resolution attention maps; and 3) a label agnostic segmentation network which generates segmentation masks given the attention map of a concept. Experiments show that our system performs competitively to state-of-the-art semantic segmentation models on concepts with full supervision, and is able to generate segmentation results for a large number of concepts with different levels of weak supervision, and even for unseen concepts.


\title{Supplementary Material}
\author{}
\institute{}
\maketitle

\section{Examples of Pseudo-Ground Truth Masks}
\begin{figure*}[ht]
\includegraphics[width=0.9\textwidth]{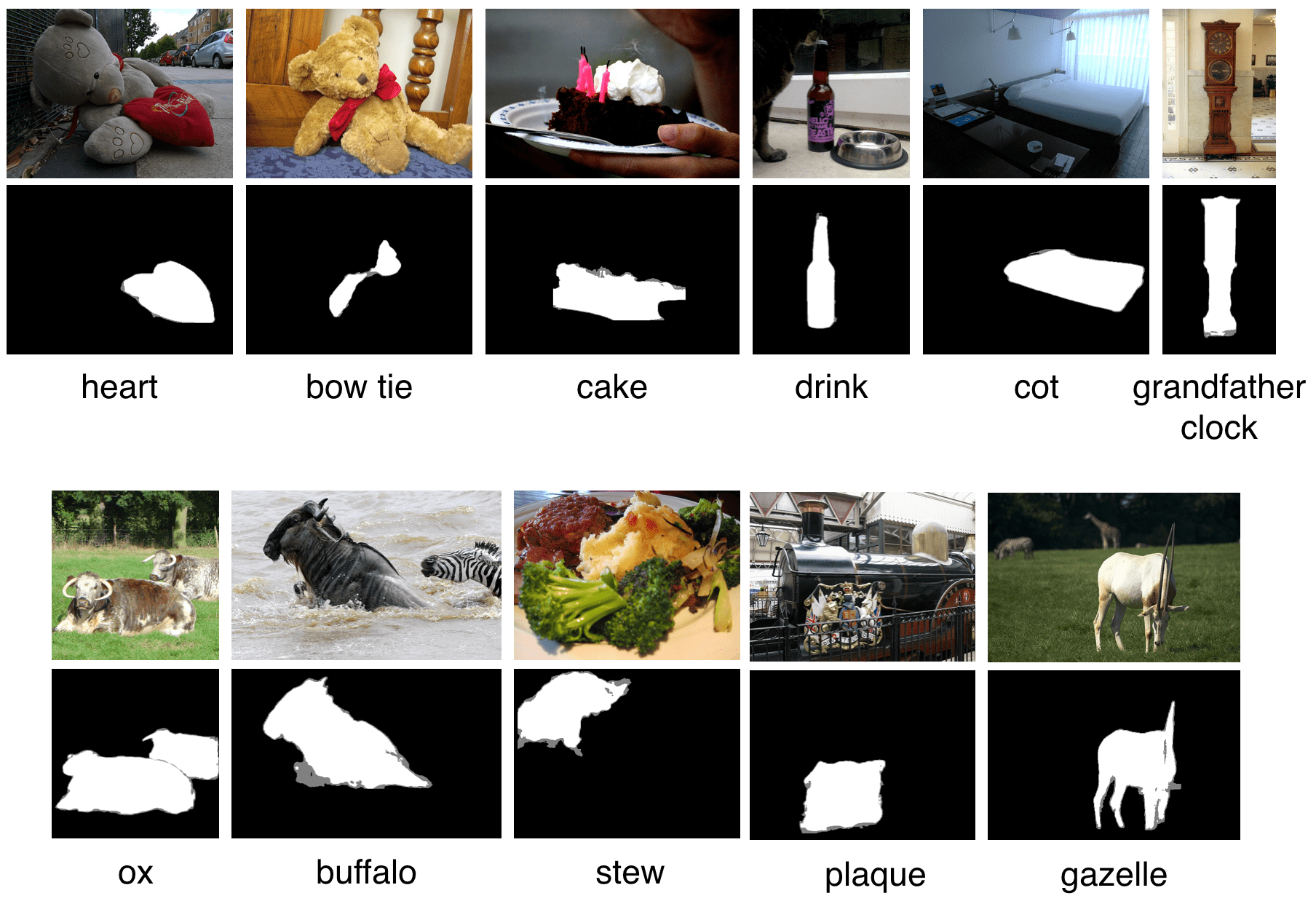}
\centering
\caption{Examples of the pseudo-groundtruth we use for evaluation on weakly-supervised settings. Note that during training, only bounding box annotations are used and such pseudo-groundtruth is not used.}
\label{gt}
\end{figure*}

In our main paper, we mention that for testing on weakly-supervised settings, there is no available segmentation ground truth for classes outside COCO-80. We therefore generate pseudo-ground truth from bounding boxes, using an automatic segmentation model \cite{xuning}. Note that we use the pseudo-ground truth masks only for evaluation, and during training only bounding boxes are used. Here in Figure~\ref{gt}, we show some examples of the generated pseudo-ground truth masks.

\section{Details of Embedding Network}
\subsection{Tag Embedding from PMI}
In the embedding network, when learning the large scale visual-semantic embedding from Stock-18K dataset, in order to calculate word embedding of a tag, we use point-wise mutual information (PMI). We calculate the PMI matrix $M$, in which the the $(i,j)$-th element is: $M_{ij}=\PMI(w_i, w_j) = \log{\frac{p(w_i, w_j)}{p(w_i)p(w_j)}}$,
where $p(w_i,w_j)$ denotes the co-occurrence probability between $w_i$ and $w_j$, and $p(w_i)$ and $p(w_j)$ denote occurrence frequency of $w_i$ and $w_j$, respectively. Matrix $M$ is of size $V \times V$, where $V$ is size of tag vocabulary $\mathbb{W}$. Apparently $M$ accounts for the co-occurrences of tags in the training corpus. Eigenvector decomposition is then applied to decompose the matrix $M$ as $M=USU^T$. Let $W=US^{\frac{1}{2}}$, then each row of the column-truncated submatrix $W_{:,1:D}$ is used as the word vector.

\subsection{Overlap Ratio Matrix for Spatial Discrimination Loss}
\begin{figure*}[ht]
\centering
\includegraphics[width=0.65\linewidth]{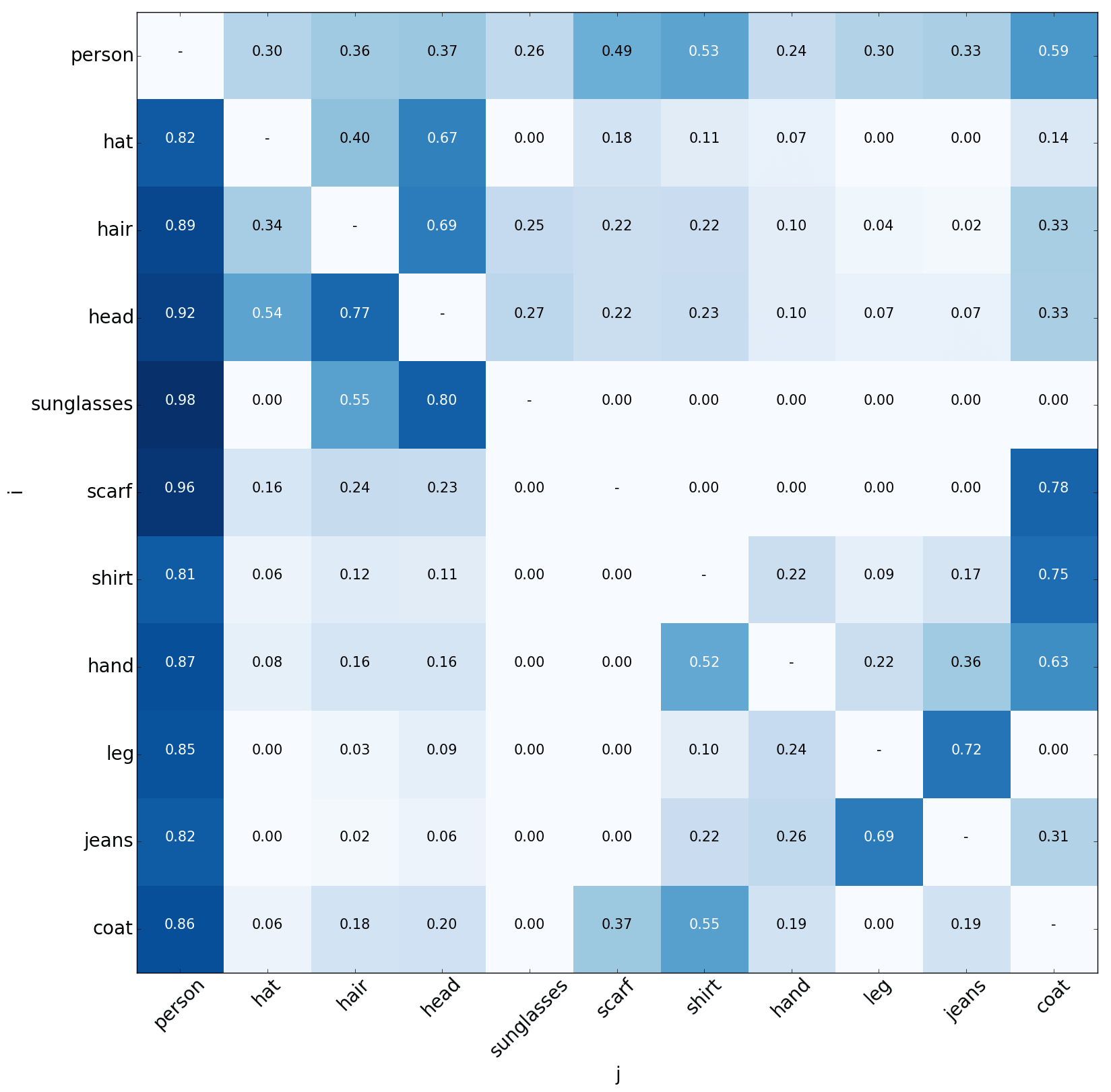}
\caption{Partial overlap ratio matrix on 11 concepts related to person.}
\label{confusion_matrix}
\end{figure*}

For attention map refinement, we introduce a spatial discrimination loss to make the training more discriminative. Here we show a subset of the overlap ratio matrix $O(i,j)$ in Figure~\ref{confusion_matrix}. 11 person-related concepts are shown here.With the vertical and horizontal axis being $i$ and $j$ respectively, we can see how different body parts overlap with person or each other. 
\clearpage

\section{Architecture of Segmentation Network}
Here in this section, we describe our Segmentation Network we used in the main paper in detail.


The proposed segmentation network is composed of three parts:
\paragraph{High-level Stream}
This part is a traditional deep CNN encoder network, except the input to the network has two extra channels of attention map we obtained in the attention network (one from input image size 300 $\times$ 300, one from size 700 $\times$ 700). For our segmentation model, we use a version of the Inception-V2 \cite{rethinking}. We remove the last three layers, i.e. Pool, Linear and Softmax in Table 1 of \cite{rethinking}. The input of our model is a 244x244 5-channel image$+$attention map and the original output of the truncated Inceptions-V2 is a 7x7 1024-channel feature map. To get a 14x14 feature map, we use dilated convolution for the last two inception modules. Finally, we add a convolution layer to generate the 2-channel 14x14 feature map. 

\paragraph{Low-level Stream}
The low-level stream is a shallow network.  The input to the shallow network is the 3-channel image and two extra channels of attention map. Specifically, we use a single 7x7 convolution layer with stride of 1. The output of this stream is a 64-channel 224x224 feature map.

\paragraph{Dense Module}
The dense module takes the low-level and high-level feature as input and outputs the final result. More specifically, we will first resize the high-level feature map to the original resolution (224x224 in our case) by bilinear upsampling. Then, we concatenate the upsampled high-level feature map with the low-level feature map and pass them to the densely connected layer units. Each dense unit is composed of some convolutional layers, and the output will be concatenated with the input to the unit.
\section{Experimental Details of Our Model}
This section describes the experimental details of our embedding network and segmentation network. 

For the embedding network, the CNN extractor we use is ResNet-50, and the three fully connected layers have dimension of 4096. The embedding vector for concepts is 4096-dimensional. During training, the input images are resized to 300$\times$300, and the size of attention map is 10$\times$10. For the refinement network, the first bottleneck convolutional layer is 2000-dimensional. The threshold used to generate binary mask for original attention mask is set to 0.2. For the segmentation network, the input image size is fixed  to 224$\times$224, and the resulting segmentation mask is resized back to the original image size.

\section{Verification of Existence of Segmented Concepts}
In our main paper, after an attention map is obtained, we normalize it to $[0, 1]$ and use the normalized attention map as input to the segmentation network. The verification of the existence of a concept can be done separately by a separate classification network. Here, although our main goal is segmentation when a concept is given, and the objective of our attention network does not include classification, we show that discrimination is achieved in our attention network.

Once we have an attention map for a given concept, our verification is done in several steps:
\begin{itemize}
\item Normalize the attention map so that it sums to 1.
\item Use the attention map as the weight to conduct a weighted average pooling on the ResNet features, rather than global average pooling in Figure 2 in the main paper.
\item The pooled feature is then passed into the embedding network, as shown in Figure 2 in the main paper. The output of the embedding network is the similarity score with the given concept to verify.
\item A separate threshold is used for each of the concept. It is decided on a validation dataset.
\end{itemize} 

Open Images and Visual Genome dataset are not fully annotated datasets, thus are not suitable for evaluation of our verification framework. Therefore, we show the effectiveness of our verification framework on the test set of COCO-80, a 5000 image dataset from COCO-validation set. Our framework achieved 70\% precision and 63\% recall on the 80 classes. It can be seen that although our attention network is not trained for classification, it can achieve decent classification result.

\begin{figure*}[t]
\includegraphics[width=\textwidth]{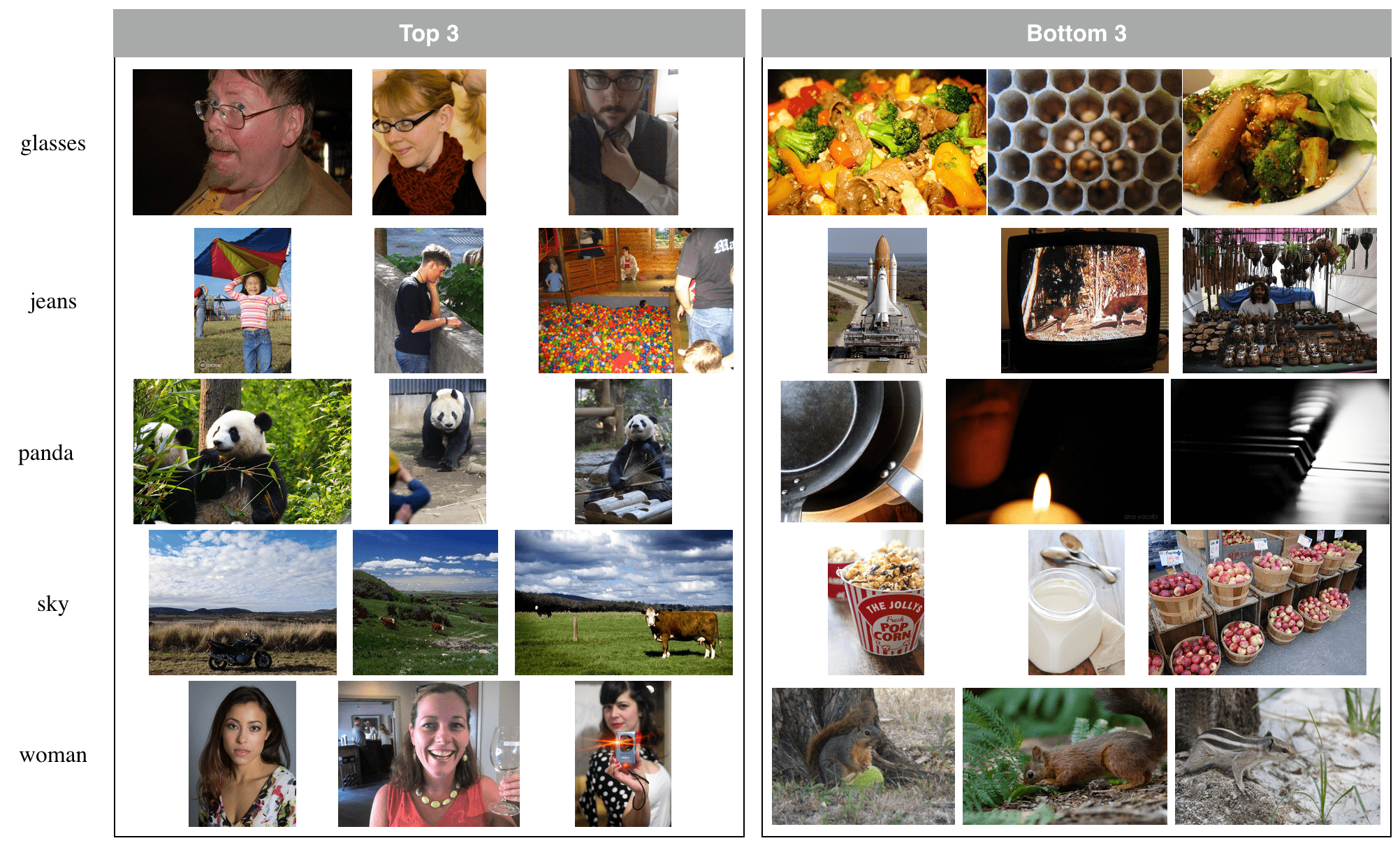}
\centering
\caption{In Weak-Box-670, top 3 images and bottom 3 images for example concepts.}
\label{top3}
\end{figure*}

\begin{figure*}[t]
\includegraphics[width=\textwidth]{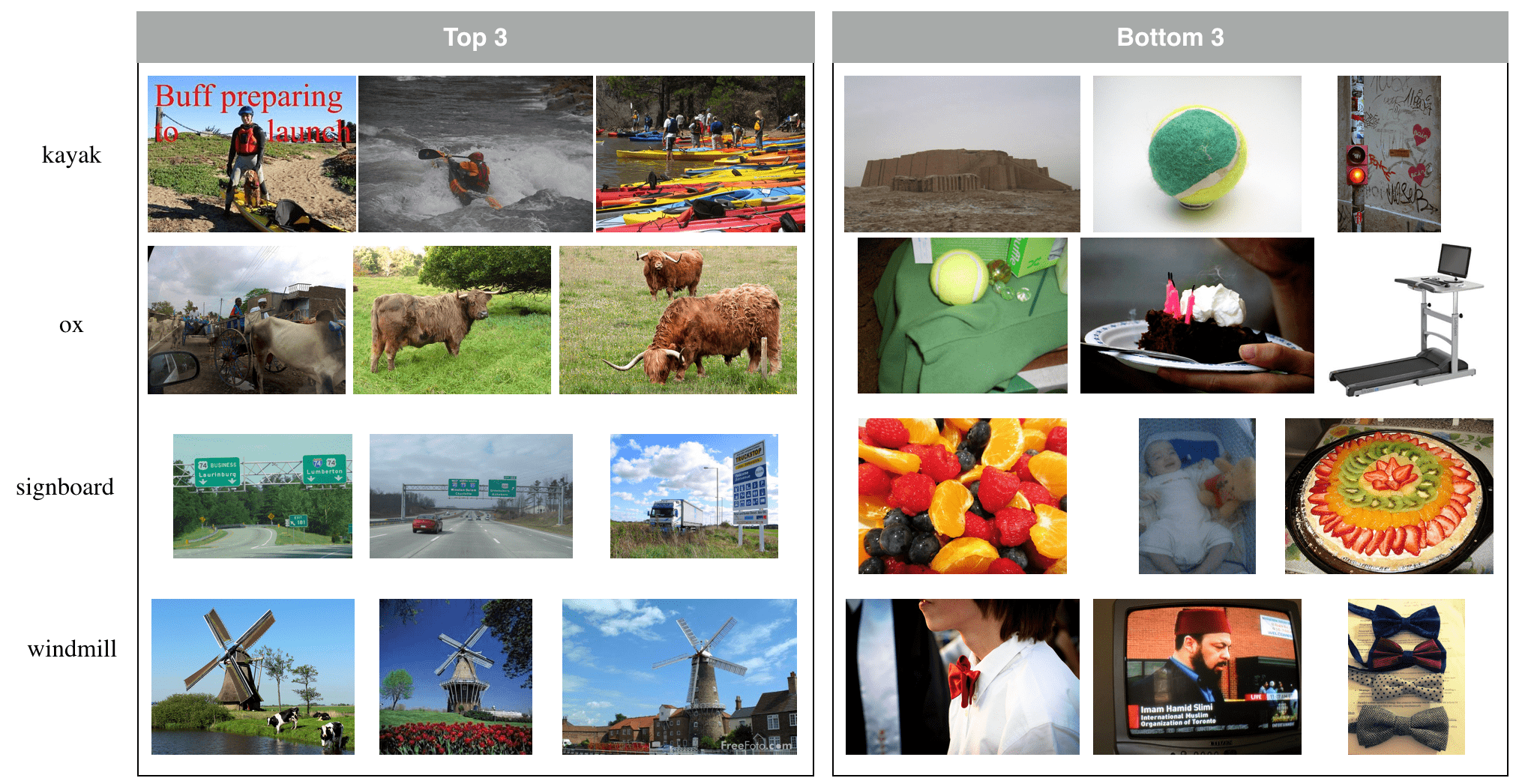}
\centering
\caption{In Weak-Image-50, top 3 images and bottom 3 images for example concepts.}
\label{top3-zero}
\end{figure*}

\begin{figure*}[t]
\includegraphics[width=\textwidth]{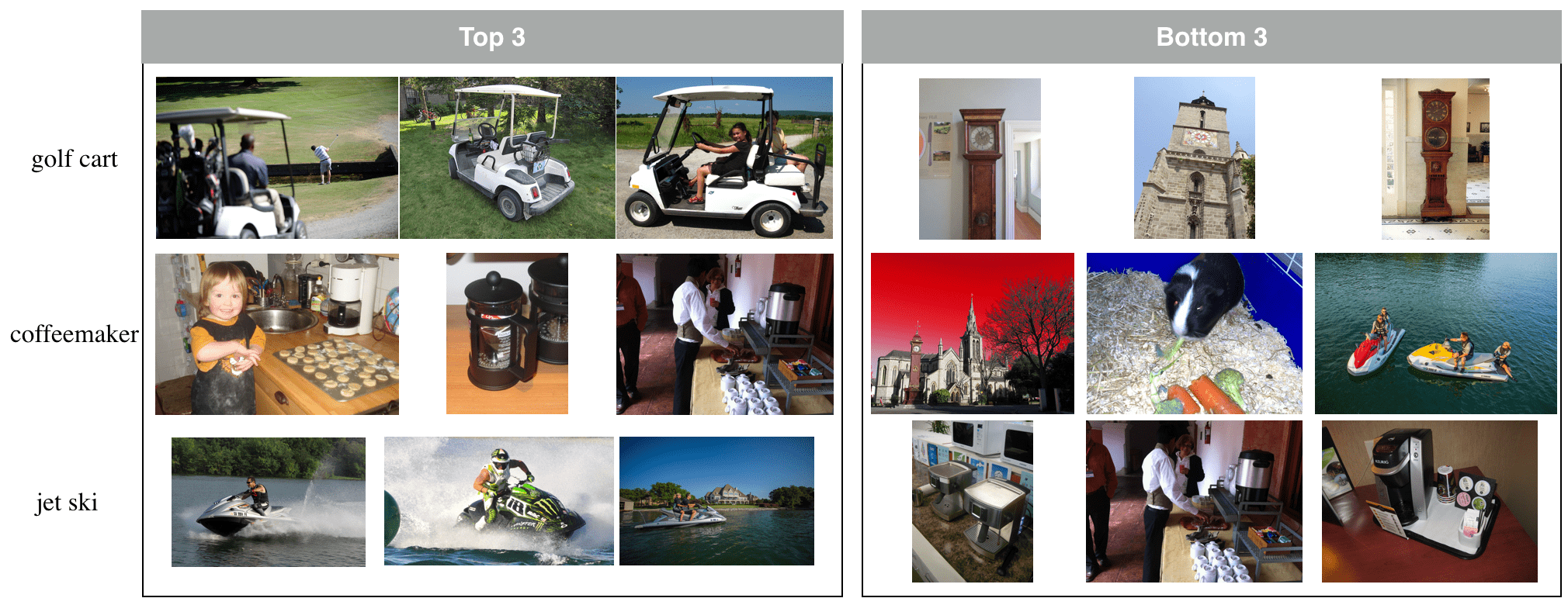}
\centering
\caption{In ZeroTest-10, top 3 images and bottom 3 images for example concepts.}
\label{top3-zero-real}
\end{figure*}

For other test sets, we can visualize the images with highest confidence scores, and the images with the lowest scores. Figure~\ref{top3} shows 5 example concepts in Weak-Box-670 and the 3 images with the highest scores, and 3 images with the lowest scores from Weak-Box-670 test set. Figure~\ref{top3-zero} shows four concepts from Weak-Image-50. In the main paper, we also mention we curate 10 concepts outside the 18K concepts that our model is trained on, each with 5-10 test examples. We call the 10-concept test set \textbf{ZeroTest-10}. Figure~\ref{top3-zero-real} shows 3 concepts from ZeroTest-10. For the three datasets, top images clearly contain the concept of interest, while bottom images do not. This also shows that our attention network has the ability for concept verification/classification.

\section{Performance for Object/Stuff/Part Classes}
We show in the main paper our model's performance with different levels of supervision. We mention in the main paper that our model works on not only object classes, but also stuff and object parts. Here, in Table~\ref{performance}, we show the performance on the object/stuff/part classes on Dataset \textbf{Weak-Box-670}. It can be seen that our model works on all the three types of classes.

\begin{table}[ht]
\centering
\caption{Our model's performance on \textbf{Weak-Box-670}, for object/stuff/part classes}
\label{my-label}
\begin{tabular}{l|lll}
Class type & Precision & Recall & IoU   \\ \hline
Object     & 0.742     & 0.791  & 0.603 \\ \hline
Stuff      & 0.713     & 0.868  & 0.625 \\ \hline
Part       & 0.709     & 0.680  & 0.516
\end{tabular}
\label{performance}
\end{table}

\section{Visual Results of Our Model}
\begin{figure*}[t]
\includegraphics[width=0.8\textwidth]{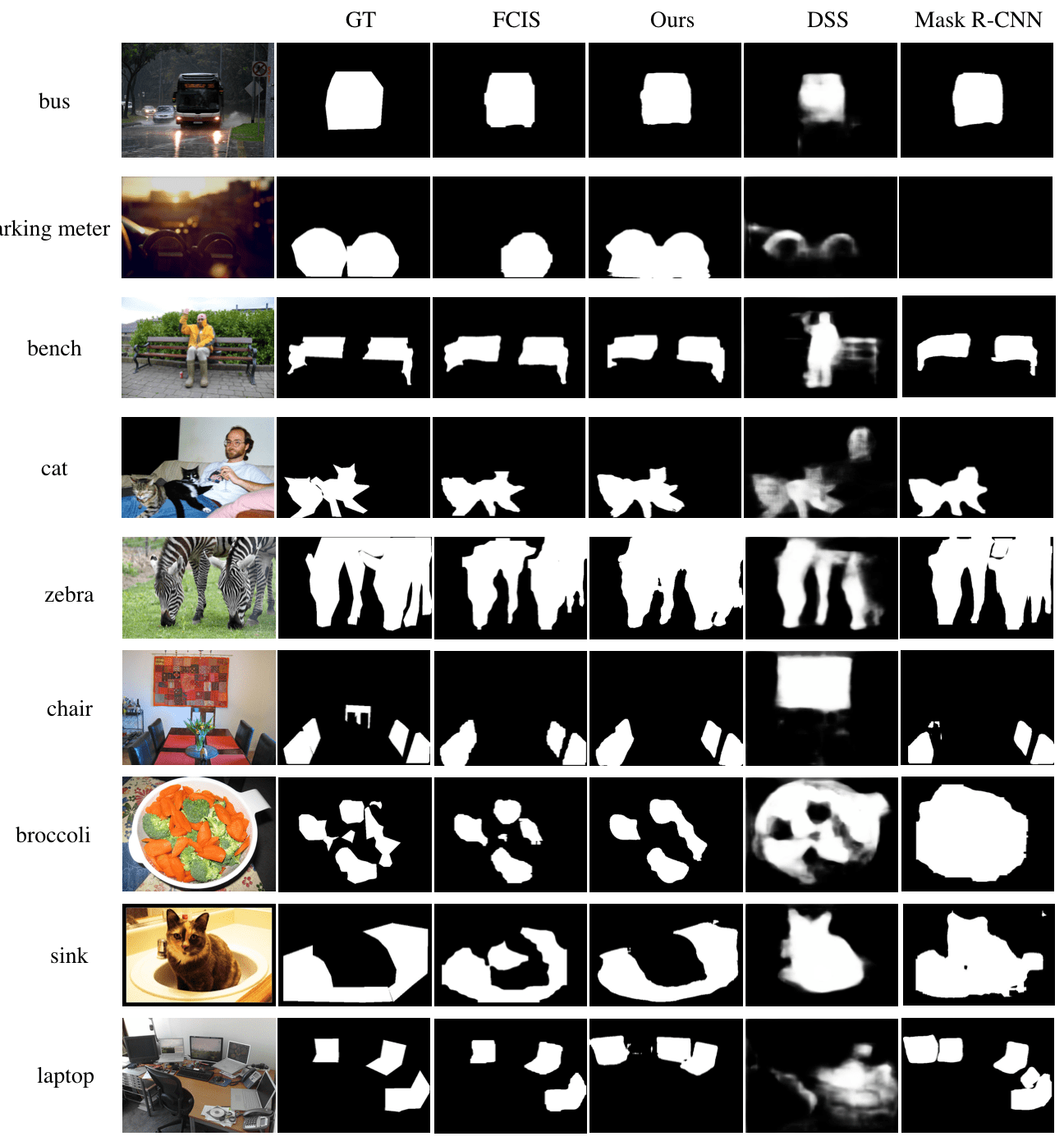}
\centering
\caption{Some examples of result on COCO-80.  Each row shows an example, and the columns show original image, groundtruth, FCIS, ours, DSS and Mask-RCNN respectively.}
\label{coco-result}
\end{figure*}

We show more visual results of our model for different test sets. Figure~\ref{coco-result} shows the examples of result on COCO-80. We compare our result with FCIS \cite{fcis} as mentioned in the main paper, the saliency algorithm DSS \cite{dss}, and our modified version of Mask R-CNN. As we can see, our result is quite close to FCIS, and some times is better (e.g. the example of parking meter). In the last row, the example of laptop shows a failure case of our algorithm, where it does not distinguish desktop with laptop perfectly. This is partially due to the training data containing Visual Genome and Open Images not annotated perfectly. 

\begin{figure*}[t]
\includegraphics[width=0.8\textwidth]{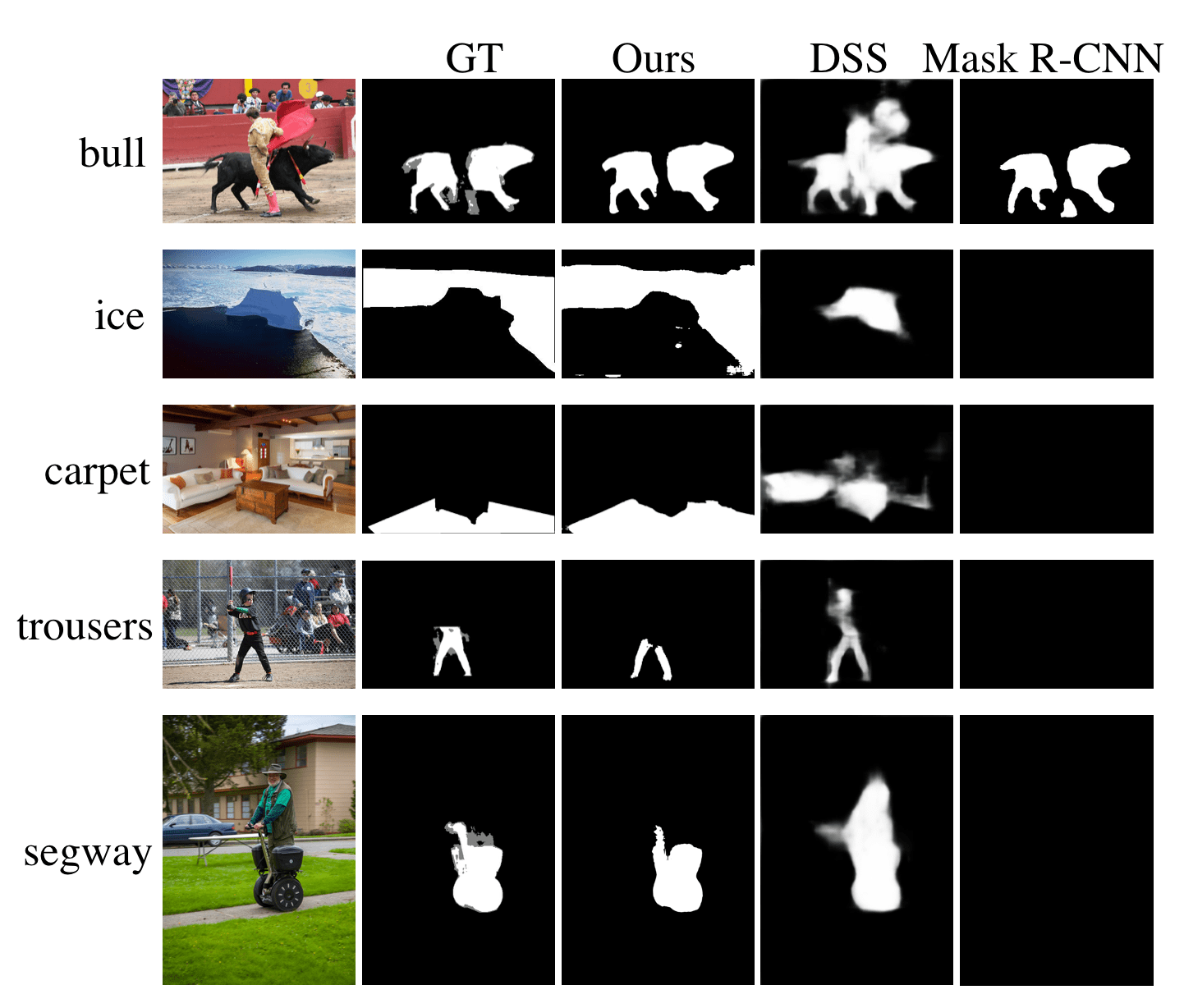}
\centering
\caption{Some examples of result on Weak-Box-670.  Each row shows an example, and the columns show original image, groundtruth, ours, DSS and Mask-RCNN respectively.}
\label{weak1}
\end{figure*}

\begin{figure*}[t]
\includegraphics[width=0.8\textwidth]{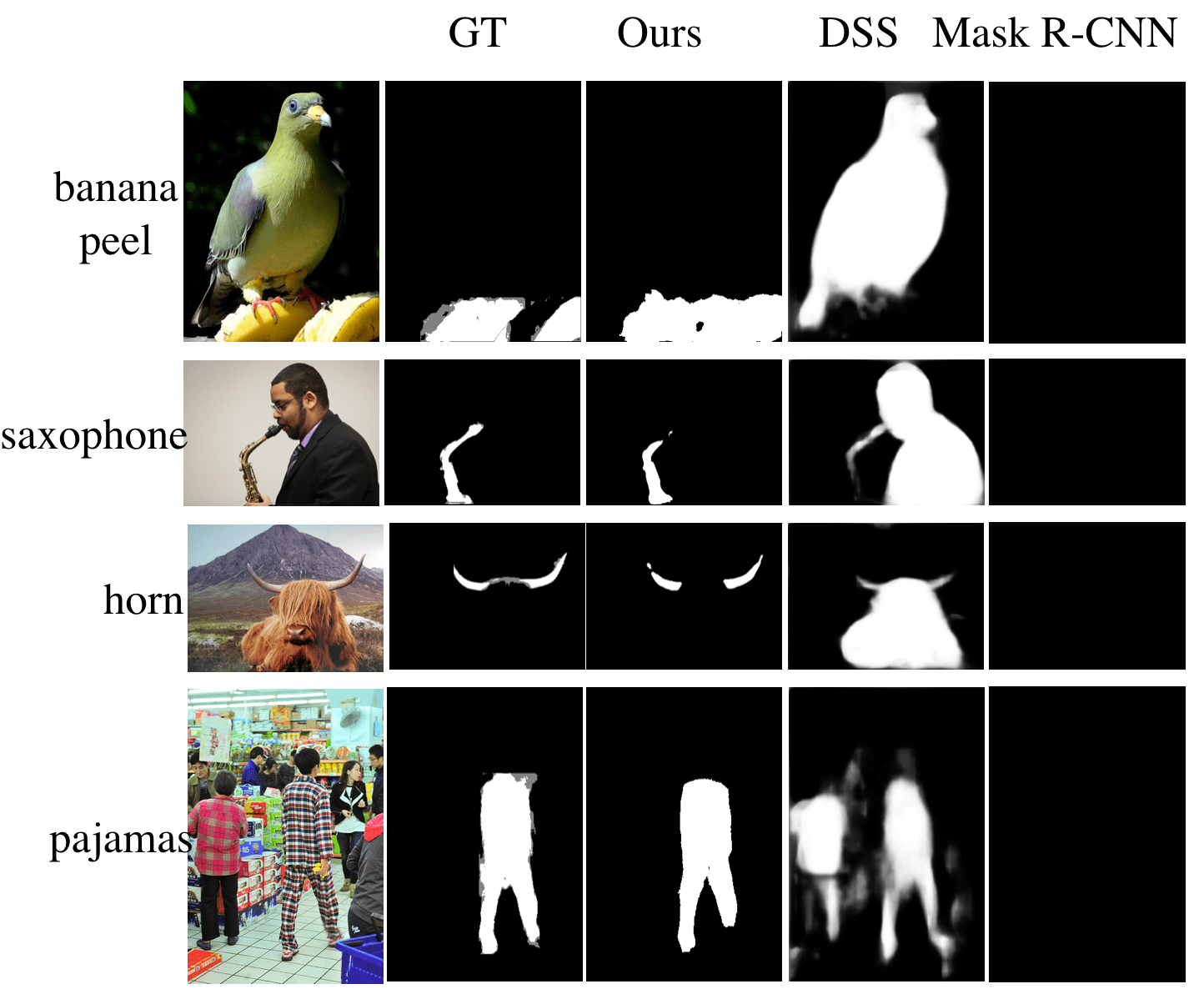}
\centering
\caption{Some examples of result on Weak-Box-670.  Each row shows an example, and the columns show original image, groundtruth, ours, DSS and Mask-RCNN respectively (continued).}
\label{weak2}
\end{figure*}

\begin{figure*}[t]
\includegraphics[width=0.8\textwidth]{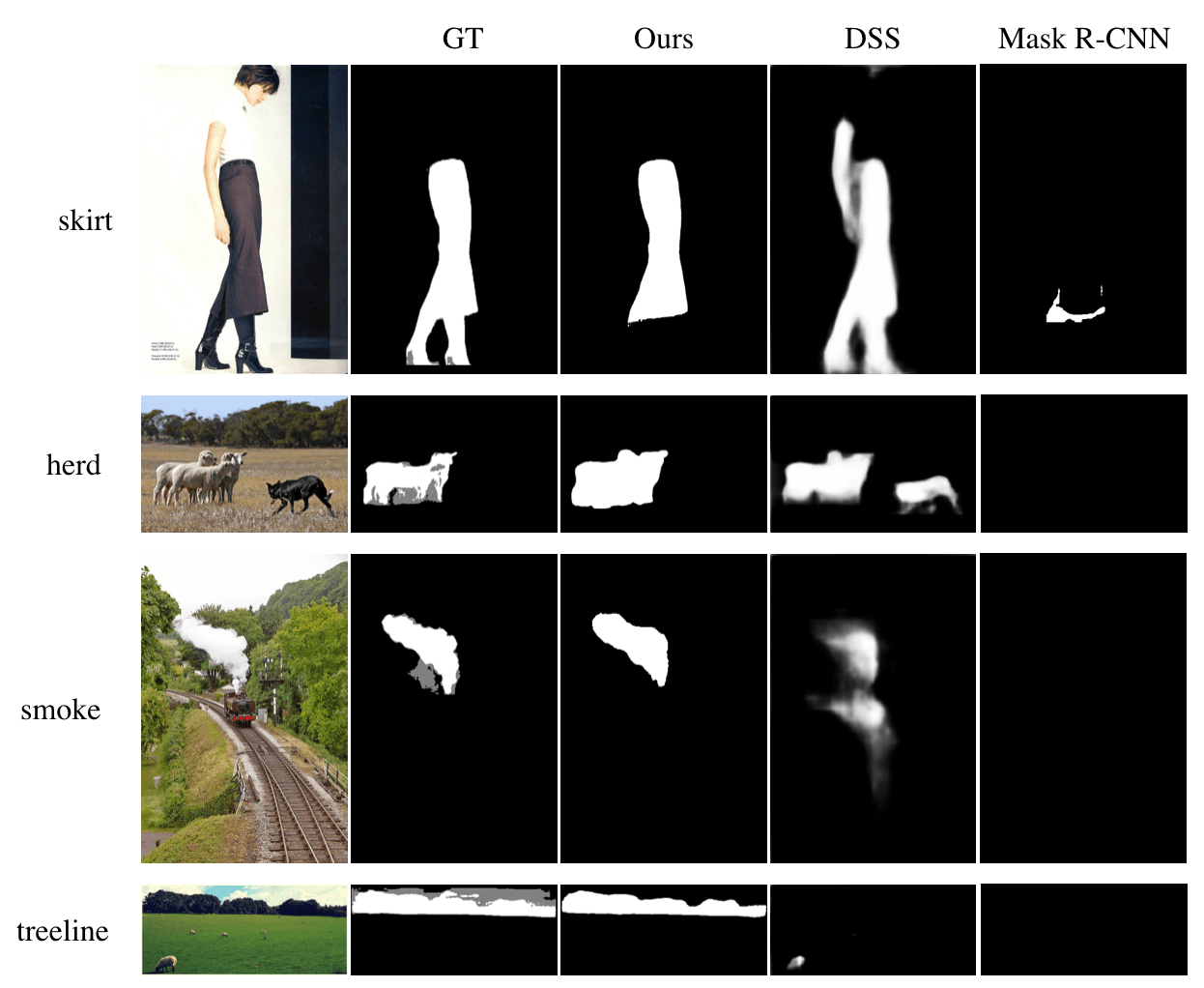}
\centering
\caption{Some examples of result on Weak-Box-670.  Each row shows an example, and the columns show original image, groundtruth, ours, DSS and Mask-RCNN respectively (continued).}
\label{weak3}
\end{figure*}

\begin{figure*}[t]
\includegraphics[width=0.8\textwidth]{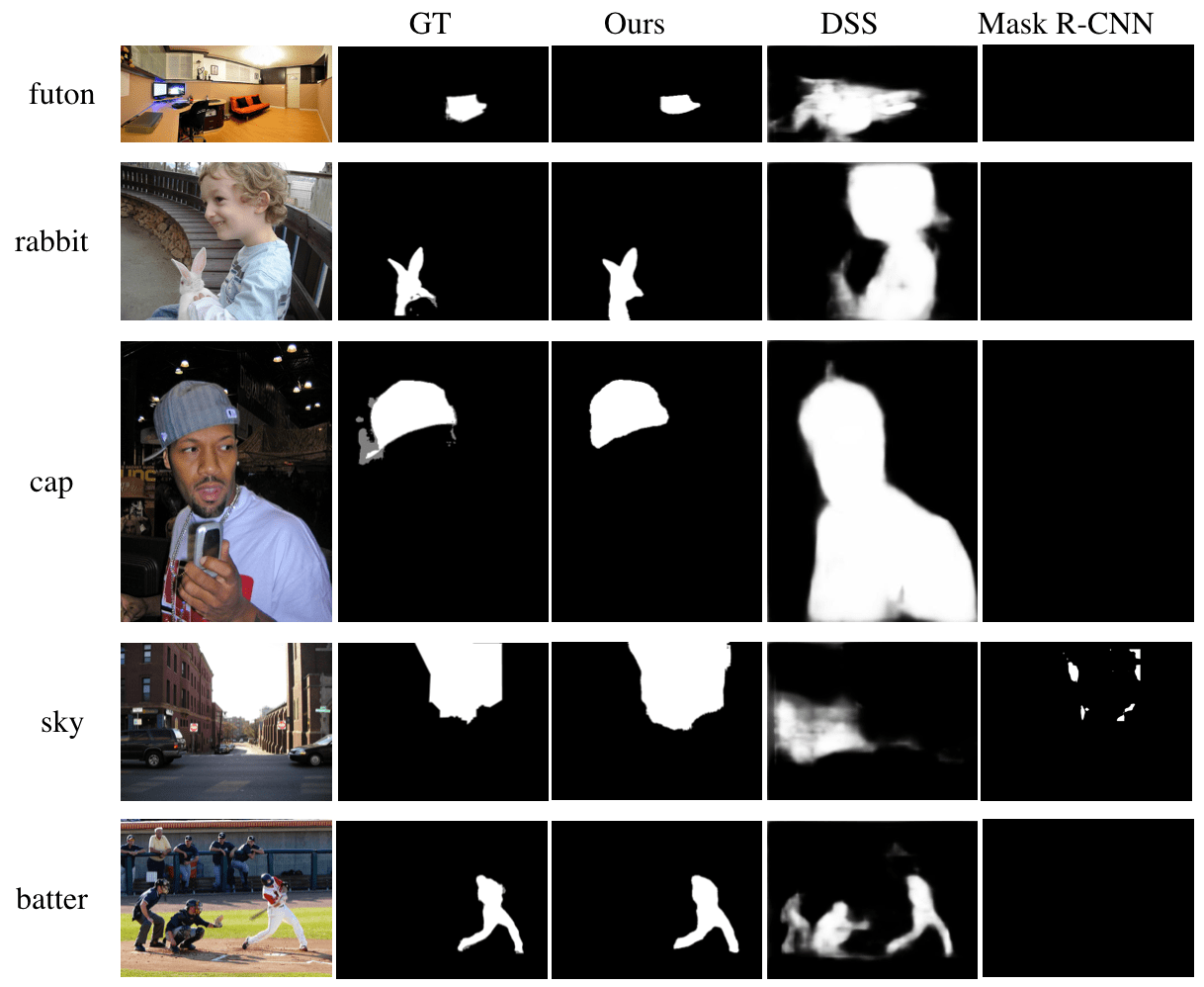}
\centering
\caption{Some examples of result on Weak-Box-670.  Each row shows an example, and the columns show original image, groundtruth, ours, DSS and Mask-RCNN respectively (continued).}
\label{weak4}
\end{figure*}

In Figure~\ref{weak1} to Figure~\ref{weak4}, we show some examples of our result on Weak-Box-670 test set. We also show DSS and modified Mask R-CNN as a baseline, to show the validity of our segmentation problem. Note that although our segmentation network never sees object parts or stuff, our algorithm has the ability to segment object parts/stuff out (e.g. trousers and horn are object parts, while treeline and sky are stuff).

\begin{figure*}[t]
\includegraphics[width=0.8\textwidth]{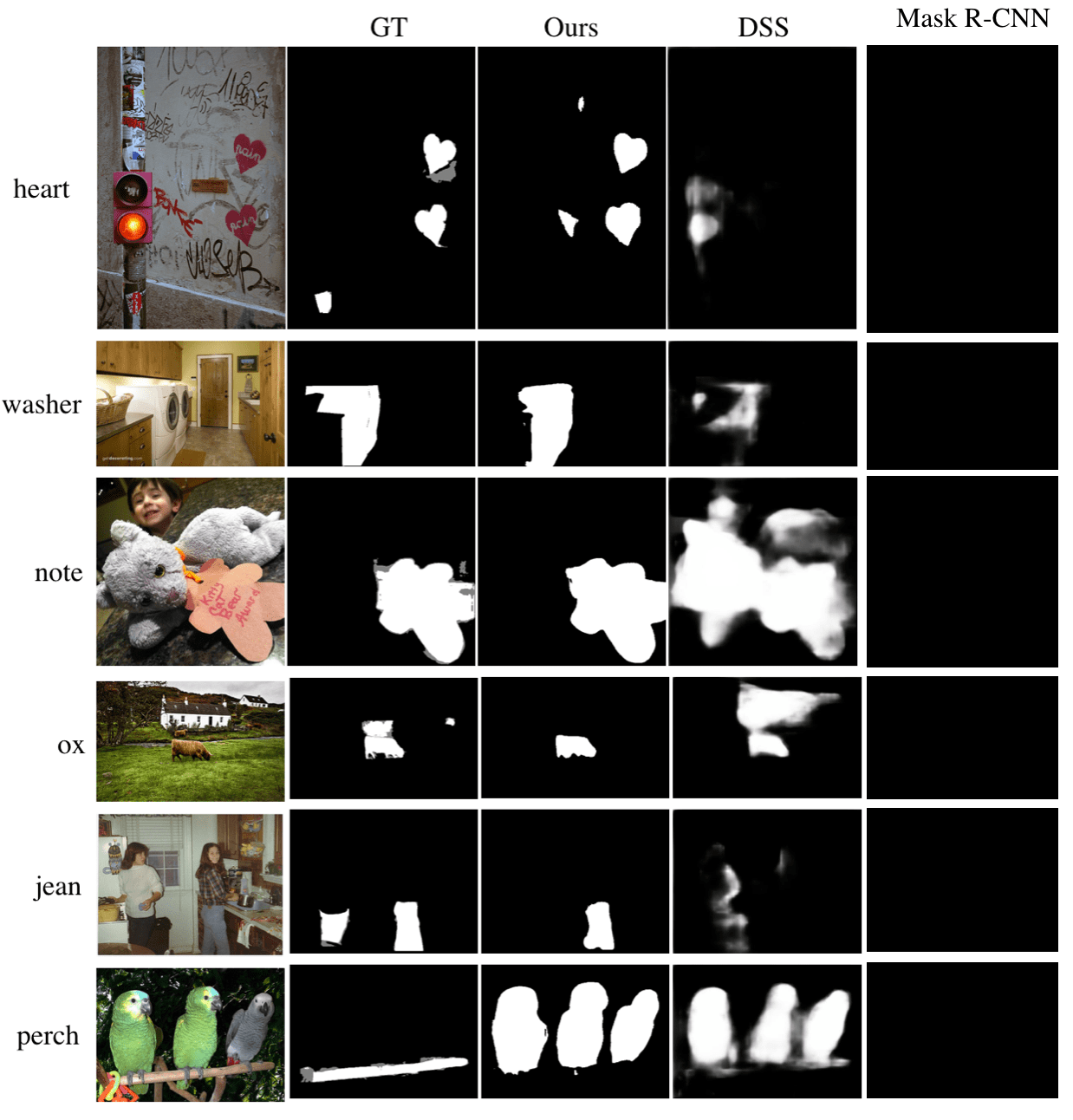}
\centering
\caption{Some examples of result on Weak-Image-50.  Each row shows an example, and the columns show original image, groundtruth, ours, DSS and Mask-RCNN respectively.}
\label{weak_img}
\end{figure*}

In Figure~\ref{weak_img}, we show some examples of our result on Weak-Image-50. The network only sees image-level labels for those categories, but our model can still segment the concept out. However, in the last row, we show a failure case, where perch always appears with birds in images, and image level labels are not enough to teach the network where the perch concept really locates at.

\begin{figure*}[t]
\includegraphics[width=0.8\textwidth]{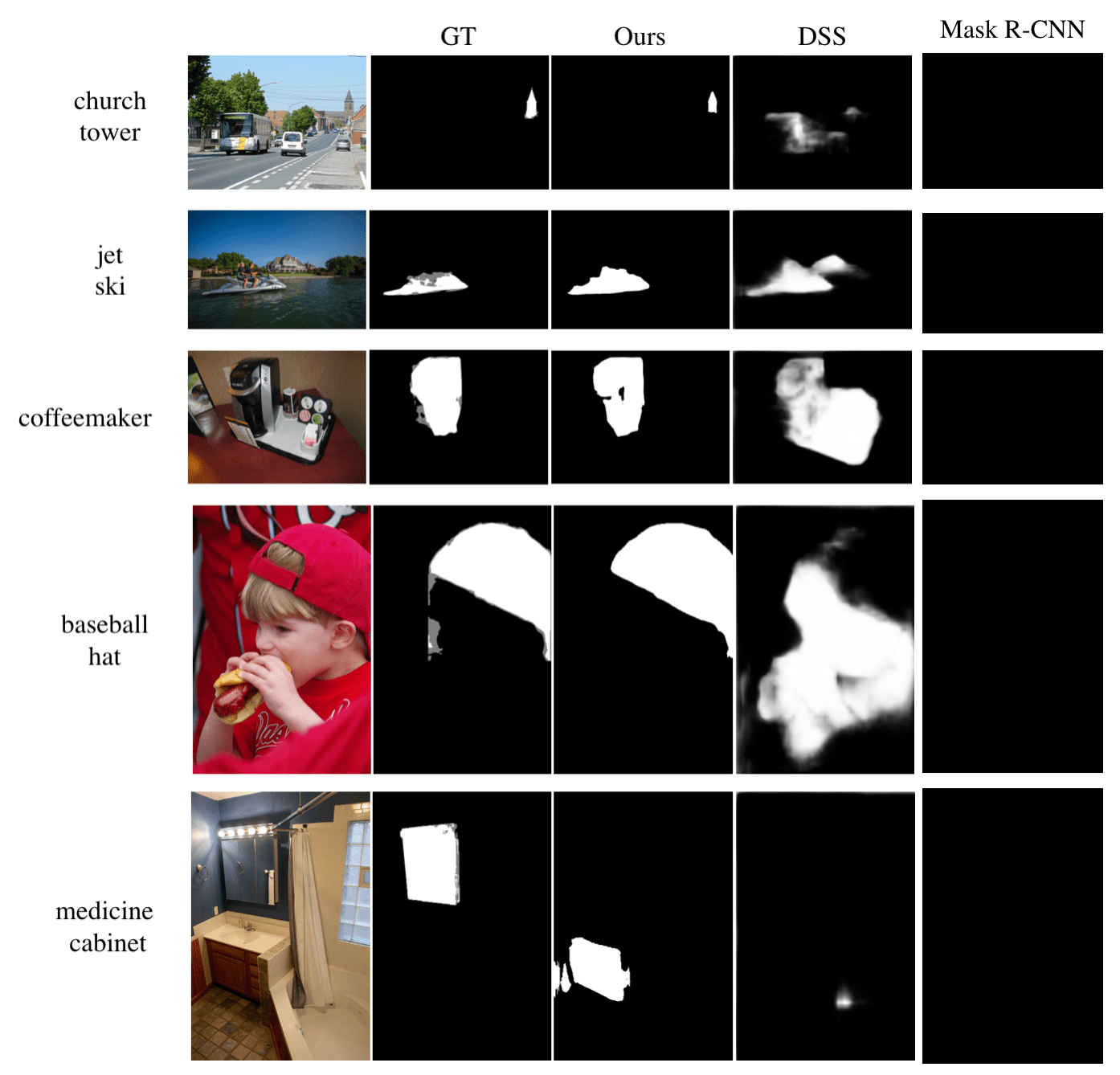}
\centering
\caption{Some examples of result on ZeroTest-10.  Each row shows an example, and the columns show original image, groundtruth, ours, DSS and Mask-RCNN respectively.}
\label{zero4}
\end{figure*}

In Figure~\ref{zero4}, we show some examples of our result on ZeroTest-10. The network shows the ability to do zero-shot learning. In the last row, we show that when the concept is easily confused with other concepts (medicine cabinet with cabinet), our model fails to segment the correct object.

\begin{figure*}[t]
\includegraphics[width=1\textwidth]{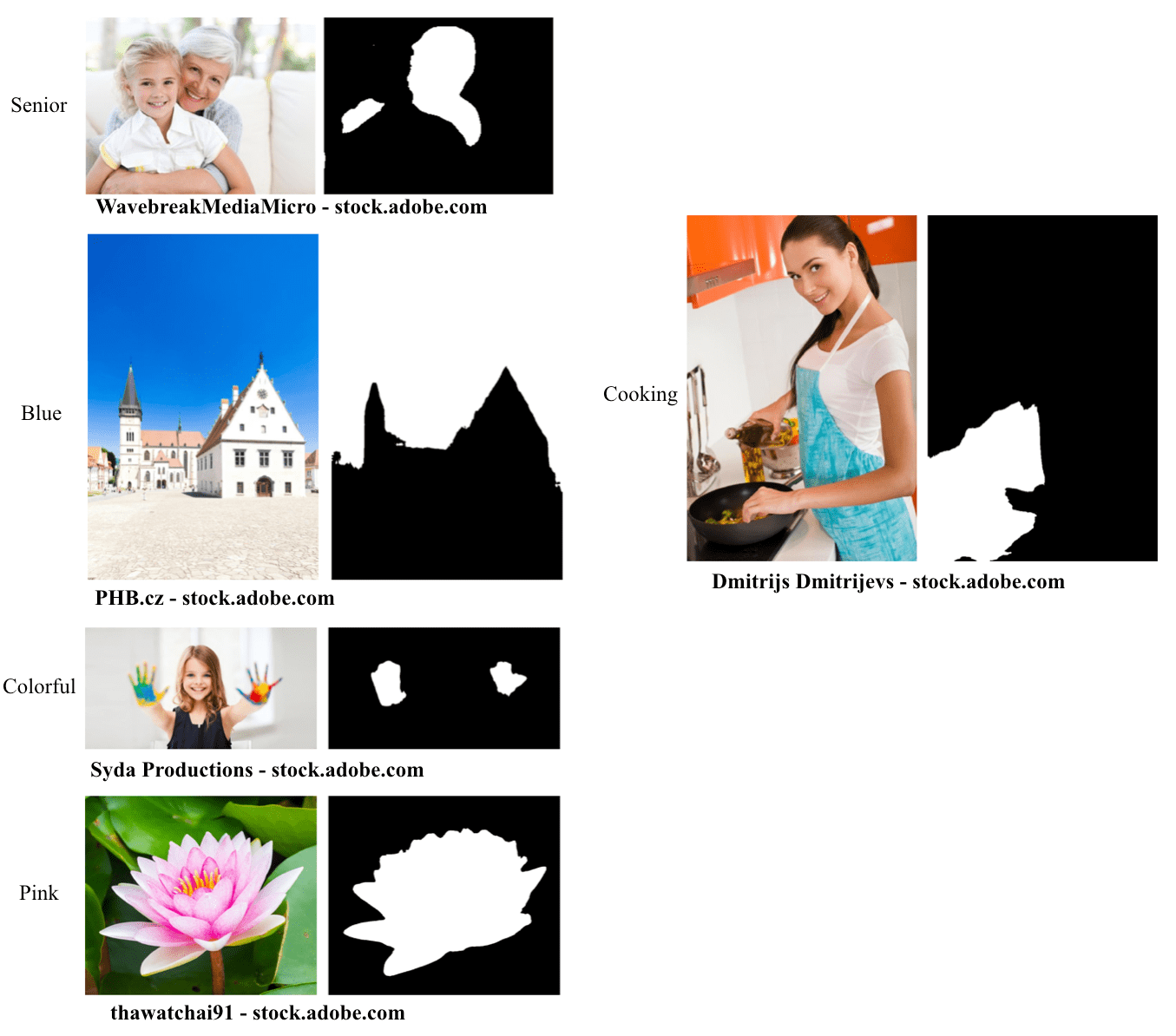}
\centering
\caption{Several examples of our model segmenting adjective/verb words. For each row, the two images are original image and segmentation mask respectively. The images are from \url{https://stock.adobe.com}, and the source is noted below each image. }
\label{adj}
\end{figure*}

In Figure~\ref{adj}, we show that our model also has the ability to segment out adjective and verb words, even though the segmentation network is only trained with objects.

\begin{figure*}[ht]
\includegraphics[width=0.5\textwidth]{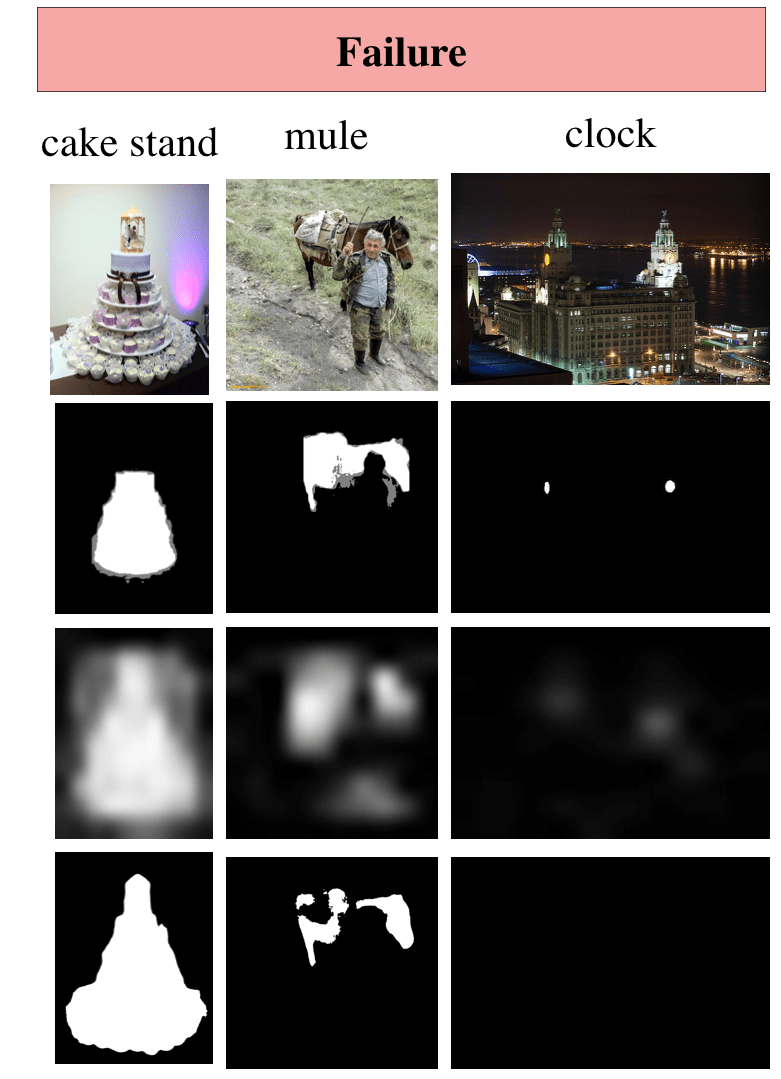}
\centering
\caption{Failure examples of attention map and mask prediction.}
\label{failure_result}
\end{figure*}

In Figure~\ref{failure_result}, we show some failure cases of our model. We also show attention map prediction along with the segmentation mask.

\clearpage

\bibliographystyle{splncs04}
\bibliography{egbib}

\end{document}